# Extracting Consumer Insight from Text:
# A Large Language Model Approach to Emotion and Evaluation Measurement

**Stephan Ludwig**
Professor of Marketing
Monash University, Australia
Email: stephan.ludwig@monash.edu

**Peter J. Danaher***
Professor of Marketing and Econometrics
Monash University, Australia
E-mail: peter.danaher@monash.edu

**Xiaohao Yang**
PhD Candidate in Data Science and AI
Monash University, Australia
Email: Xiaohao.Yang@monash.edu

**Yu-Ting Lin**
Senior Lecturer in Marketing
University of New South Wales, Australia
Email: yuting.lin1@unsw.edu.au

**Ehsan Abedin**
Lecturer in Business Analytics
Flinders University, Australia
Email: ehsan.abedin@flinders.edu.au

**Dhruv Grewal**
Toyota Professor of Commerce and Electronic Business
Marketing Division, Babson College
213 Malloy Hall, Babson Park, MA 02457, USA
Email: dgrewal@babson.edu
Fractional Professor, University of Bath, UK
Honorary Distinguished Visiting Professor of Retailing and Marketing,
Tecnológico de Monterrey, Monterrey, Mexico
ORCiD: https://orcid.org/0000-0002-7046-6063

**Lan Du**
Associate Professor in Data Science and AI
Monash University, Australia
Email: Lan.Du@monash.edu

Acknowledgments: The authors thank the Australian Research Council (ARC) for partial financial support during this project (ARC Discovery Project: DP230101490).

*Corresponding author.

# Extracting Consumer Insight from Text:
## A Large Language Model Approach to Emotion and Evaluation Measurement

**Abstract**

Accurately measuring consumer emotions and evaluations from unstructured text remains a core challenge for marketing research and practice. This study introduces the Linguistic eXtractor (LX), a fine-tuned, large language model trained on consumer-authored text that also has been labeled with consumers' self-reported ratings of 16 consumption-related emotions and four evaluation constructs: trust, commitment, recommendation, and sentiment. LX consistently outperforms leading models, including GPT-4 Turbo, RoBERTa, and DeepSeek, achieving 81% macro-F1 accuracy on open-ended survey responses and greater than 95% accuracy on third-party–annotated Amazon and Yelp reviews. An application of LX to online retail data, using seemingly unrelated regression, affirms that review-expressed emotions predict product ratings, which in turn predict purchase behavior. Most emotional effects are mediated by product ratings, though some emotions, such as discontent and peacefulness, influence purchase directly, indicating that emotional tone provides meaningful signals beyond star ratings. To support its use, a no-code, cost-free, LX web application is available, enabling scalable analyses of consumer-authored text. In establishing a new methodological foundation for consumer perception measurement, this research demonstrates new methods for leveraging large language models to advance marketing research and practice, thereby achieving validated detection of marketing constructs from consumer data.

# 1. Introduction

Accurately measuring consumer perceptions at scale, especially those expressed through emotions and evaluations, remains a central challenge for marketing research. Perceptions linked to emotions such as anger, anxiety, and joy or to evaluations like trust and commitment deeply inform consumers' decisions and firm outcomes (Cohen, Pham, and Andrade 2018; Gopinath, Thomas, and Krishnamurthi 2014; Gupta and Zeithaml 2006; Kranzbühler et al. 2020; Morgan and Hunt 1994; Palmatier et al. 2006). To identify such perceptions at scale, consumer-authored text, such as reviews and social media posts, provide extremely useful data (Babić Rosario et al. 2016). Yet most text mining methods applied to consumer reviews and posts seek to reduce this vast content to simplistic, coarse (positive versus negative) sentiment labels, offering limited insight into consumers' different emotions and evaluations. Consequently, the richness of unstructured consumer data is severely underutilized. The need for methods that deliver valid, fine-grained, consumer perception metrics from unstructured data sources is urgent (Kim, Lee, and McCulloch 2024).

Marketing research has largely tended to rely on text mining methods, that are not able to capture the full breadth and depth of how consumers express emotions and evaluations. For example, lexicon-based approaches such as LIWC (Pennebaker et al. 2015), the Evaluative Lexicon 2.0 (Rocklage, Rucker, and Nordgren 2018), metaphors (Luri, Schau, and Ghosh 2023), natural affect detection (Hotz-Behofsits, Wlomert, and Nabout 2025), and VADER (Hutto and Gilbert 2014) can detect affect or sentiment at a broad level but cannot capture more granular constructs such as discontent, excitement, love, trust, or commitment. Transformer models, such as BERT and RoBERTa (Chintagunta, Hanssens, and Hauser 2016; Lu and Kannan 2025; Netzer et al. 2012; Puranam, Kadiyali, and Narayan 2021), offer more linguistic precision, yet they have not been developed to detect specific marketing constructs, validated using established marketing scales (e.g., three-item trust scale from Morgan and Hunt 1994), or benchmarked against consumers' self-reports.

Recent developments in large language models (LLMs), such as GPT, Llama 3, and DeepSeek, promise new possibilities for extracting insight from unstructured data, but their applications mainly have focused on generative tasks, such as visual product design (Sisodia, Burnap, and Kumar 2024), synthetic environments (Yang et al. 2023), or data generation (Arora, Chakraborty, and Nishimura 2024), not the validated detection of marketing relevant constructs and consumer perceptions. We know of no prior research that has fine-tuned LLMs to extract consumer emotions or evaluations, in an empirically valid way based on established marketing scales. This gap represents a methodological challenge; it also presents an opportunity to derive novel insights from consumer text data.

To address both the gap and the opportunities, we develop the Linguistic eXtractor (LX), a fine-tuned version of the Llama 3.2-3B Instruct LLM that has been trained on consumer self-reports (see the Practical Implications section later in the paper). It is designed specifically to detect two types of marketing-relevant consumer perceptions: discrete emotions (e.g., joy, discontent, fear) and evaluations (e.g., trust, commitment, recommendation, sentiment). Both categories of perceptions are foundational to understanding consumer behavior and relationships (Morgan and Hunt 1994; Palmatier et al. 2006), in that emotions capture the affective dimension of consumption, while evaluations reflect more deliberative assessments. As a novel feature of our fine-tuning method, we train LX on consumer-authored text directly, paired with those authors' responses to established marketing scales measuring their perceptions, to derive training data without requiring input from third-party coders. As a result, we show, LX more accurately identifies nuanced consumer perceptions that tend to be overlooked by lexicon-based tools or general-purpose LLMs.

To establish the viability of LX, we start by testing whether it can detect self-reported perceptions from open-ended survey data more accurately than existing models, such as GPT-4, DeepSeek, or RoBERTa. With Study 1, we also present a novel method for fine-tuning an open-source LLM (for this study, Meta AI's Llama 3.2-3B Instruct), which we designed specifically to identify 16 consumption-related emotions (Richins 1997) and four evaluations (e.g., trust, commitment, sentiment, recommendation). Whereas traditional methods rely on third-party online reviews labeled by human coders (Khan et al. 2025), we leverage survey input to align consumer-authored text with their own scaled responses. Specifically, with a large-sample, open-ended survey of 2,523 consumers from the United States, Canada, United Kingdom, and Australia, we obtained reports of recent product or service experiences from consumers who also rated their emotions and evaluations using established marketing scales (Morgan and Hunt 1994; Palmatier et al. 2006; Richins 1997). A performance evaluation of LX's prediction includes comparative assessments with transformer models (BERT, RoBERTa), state-of-the-art LLMs (GPT-3.5 Turbo, GPT-4 Turbo, and DeepSeek), and pretrained but not fine-tuned versions of Llama 3.2. The results confirm that LX outperforms the competing models in detecting 13 of the 16 emotions and all of the evaluations. Its macro-F1 score of 81% exceeds the 69% for GPT-3.5-Turbo and 68% for GPT-4 Turbo, for example. Considering LX's fewer parameters and cost-free implementation, these results are promising.

In Study 2, as a test of its generalizability, we check if LX outperforms existing models when applied to third-party–annotated customer review data from Amazon and Yelp. Trained human coders labeled the data sets, involving 3,710 Amazon and 2,193 Yelp online reviews, for the presence and polarity of marketing-relevant perceptions. These were used to further fine tune LX. Applied to separate holdout samples of 4,131 Amazon and 1,258 Yelp reviews, LX achieves macro-F1 scores in excess of 95%, substantially greater than all other models.

Study 3 offers nomological validity by presenting an application to customer clickstream and purchase data from a large U.K. online retailer, which we obtained from Wharton Customer Analytics. With this study, we seek to determine if the different consumer emotions relayed in customer reviews can predict downstream, business-relevant outcomes. This data set includes detailed records of product page views, review reading behavior, and transactions linked to more than 10,000 products. To model how LX-detected emotions in customer reviews might be linked to product ratings and, through those ratings, consumer purchase decisions, we use a system of seemingly unrelated regression equations with product-level controls. The results reveal that 12 of the 16 emotional effects are mediated by product ratings, but some emotions (e.g., discontent, peacefulness) are associated with purchases directly. Thus, adding to ongoing research on consumer emotions (Bagozzi, Gopinath, and Nyer 1999; Puccinelli et al. 2016), we demonstrate that consumer emotions can provide meaningful signals, beyond product ratings.

Finally, to support its broad adoption, we deploy and demonstrate LX as a no-code, cost-free, web application that allows users to upload consumer-authored text in standard formats (e.g., *csv) and receive outputs that indicate the presence of 16 consumption-related emotions and key evaluative indicators (e.g., trust, commitment). This tool can improve the accuracy and scalability of consumer perception measures, extract new insights from unstructured consumer text data, and democratize access to advanced text mining methods.

## 2. Consumer Perception Measurement Using LLMs

Consumer emotions and evaluations are central to marketing research in various fields, such as relationship marketing (Morgan and Hunt 1994), customer loyalty and equity (Gupta and Zeithaml 2006; Palmatier et al. 2006), and affective decision-making (Cohen, Pham, and Andrade 2018; Kranzbühler et al. 2020). A *consumer emotion* is defined as "a valanced affective reaction to perceptions of situations" (Richins 1997, p. 127; see also Ortony, Clore, and Foss 1987), which can influence both reactions and behaviors. Using the term consumer evaluations, we refer to consumers' assessment and opinion of a firm or offering, such as trust, commitment, recommendation, or satisfaction (with its contemporary equivalent being sentiment), that similarly influence consumer reactions and behaviors (Gupta and Zeithaml 2006; Morgan and Hunt 1994; Palmatier et al. 2006). A comprehensive review of six leading marketing journals[1] from 2000 to 2024, seeking any articles mentioning these terms in their titles, abstracts, or keywords, affirms the strong interest in such consumer perceptions, largely consistent over time (Figure 1). Notably, 14% (1,049 articles) of all published articles during this time period in these six journals mention at least one of these perceptions, with sentiment/satisfaction (43.6%) and emotions (17.2%) being the most frequently studied (for details, see Web Appendix A).

**Figure 1.** Research into Consumer Emotions and Evaluations

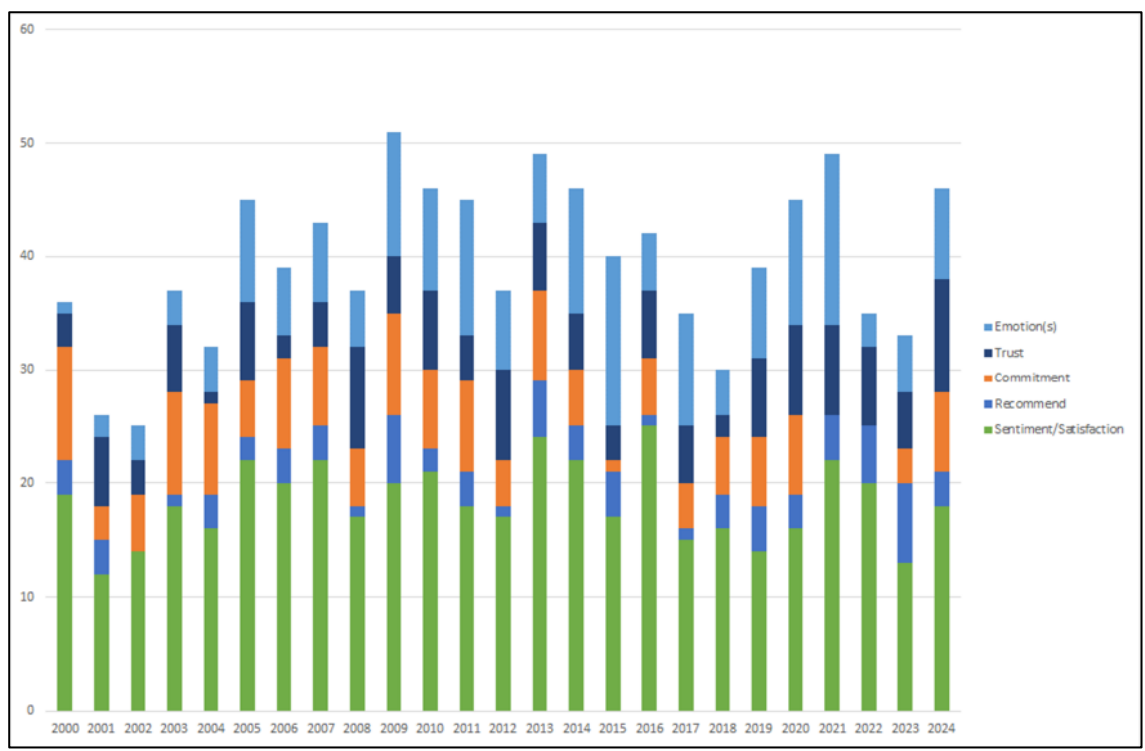

---

[1] *Journal of Marketing Research, Marketing Science, Journal of Marketing, Journal of Consumer Research, Journal of the Academy of Marketing Science*, and *International Journal of Research in Marketing*.

Yet across these substantial research streams, most studies continue to measure perceptions using survey-based scales, which may be well-validated but also are constrained in their scalability and context. The widespread availability of consumer-authored text, such as product reviews and social media posts, offers an alternative and rich source of data for revealing these perceptions, at scale, but it remains underutilized. Some marketing researchers have sought access by undertaking automated text analyses, but thus far, the available tools remain unable to deliver fine-grained measures purposefully developed to uncover specific marketing constructs. Dictionary-based tools (Hutto and Gilbert 2014; Pennebaker et al. 2015; Rocklage et al. 2018) rely on fixed lexicons to detect, for example, a few, psychologically relevant positive or negative affect categories; they cannot isolate discrete consumer emotions or evaluations like discontent, excitement, or trust. Transformer models, such as BERT and RoBERTa (Chintagunta, Hanssens, and Hauser 2016; Netzer et al. 2012; Puranam, Kadiyali, and Narayan 2021), offer greater linguistic flexibility, but they have not been adapted to detect marketing constructs, nor validated against self-reported consumer perceptions.

To understand how these constructs have been text-mined in prior work, we conducted a second review of the same six journals, seeking studies that describe using text analyses, topic modeling, or natural language processing (Web Appendix A, Figure A2). The 115 articles we found were mostly published after 2018, and only 27 of them examined consumer emotions or evaluations: 14 focused on sentiment, 4 on a handful of emotions, 8 on both, and just 1 used text mining to detect recommendations (Table A2, Web Appendix A). This narrow focus on sentiment excludes considerations of the broader, more nuanced constructs that marketing theory tends to emphasize.

Furthermore, these studies mostly leverage existing dictionary-based tools (n = 16) or neural networks with deep learning techniques (n = 4). None apply LLMs. This methodological constraint has significant theoretical implications. Without scalable, theory-aligned tools that can extract emotions and evaluations from consumer-authored text, researchers cannot effectively test or extend core marketing models to large-scale, real-world settings. This constraint also might be why existing work rarely integrates textual insights with behavioral outcomes, such that we continue to wonder if textually communicated consumer perceptions can predict marketing-relevant outcomes, such as product ratings or purchase behavior.

Furthermore, while consumer emotions and evaluations are frequently modeled as predictors of consumer choice, satisfaction, or brand engagement, existing studies often rely on data collected in small-scale, controlled environments. Many foundational studies draw on self-reports from 100–300 participants (e.g., Bagozzi, Gopinath, and Nyer 1999; Palmatier et al. 2006; Puccinelli et al. 2016), which are useful for theory testing but limited in behavioral realism. A text mining method that allows for the validated measurement of consumer emotions and evaluations can be operationalized across tens of thousands of actual text-based communication and conversation instances, helping to bridge the gap between internal validity and ecological validity for emotion research in marketing. Existing LLMs such as GPT, Llama 3, and DeepSeek offer great possibilities for extracting insights from unstructured text data sources, but they have not been fine-tuned or validated to measure marketing-relevant constructs such as consumer emotions and evaluations perceptions.

## 3. Overview of Studies

To address these critical gaps, we develop the Linguistic eXtractor (LX), a fine-tuned large language model trained to detect marketing-relevant consumer emotions and evaluations from unstructured text. To test its applicability, we conduct three empirical studies. Together, the studies demonstrate the theoretical applicability, practical utility, and behavioral relevance of the click-based LX web app, as a new method for extracting fine-grained, theory-aligned consumer perception metrics from unstructured text data sources.

Despite the growing use of large language models (LLMs) in marketing research, there remains no standardized framework for developing and validating fine-tuned models that measure core marketing constructs from unstructured text. To address this shortcoming, Table 1 offers a generalizable, step-by-step procedural guide to support future researchers in training and validating LLMs for this purpose. Although the roadmap reflects our own empirical implementation, it is designed as a flexible blueprint potentially applicable to the future development of LLM based measures of a wide range of relevant marketing constructs, including brand trust, perceived value, loyalty, service quality, authenticity, or sustainability perceptions. Each step outlines its purpose, required activities, and methodological standards to ensure construct validity, generalizability, and theoretical alignment.

The three empirical studies that follow integrate this framework. Study 1 addresses construct specification, self-report-based labelling, and model fine-tuning. Study 2 evaluates generalizability by testing performance on third-party annotated review data. Study 3 examines nomological validity by assessing whether LLM-derived outputs relate to real-world outcomes such as product ratings and purchase behavior. Together, these studies also demonstrate how the steps in Table 1 can guide researchers in developing and validating LLMs for construct-valid, scalable measurement in marketing contexts.

## 4. Study 1. A Survey-Based Approach to Training LLMs

With Study 1, we develop and fine-tune a novel LLM to detect marketing-relevant consumer perceptions from text. To support this theory-consistent training, we collect a large-scale data set that links consumer-authored text to self-reported emotions and evaluations.

**Table 1.** Procedure for Developing and Validating Fine-Tuned LLMs for Detecting Marketing Constructs from Unstructured Text

| Step | | Purpose | Core Activities | Key Recommendations and Standards |
|---|---|---|---|---|
| 1 | Construct Specification | Identify theory-grounded constructs from validated marketing literature | Define constructs from foundational literature using validated scales and identify their semantic markers in text | Use multi-item scales with α > .70 (e.g., Richins' (1997) consumer emotions; Morgan and Hunt's (1994) trust). Specify the theoretical definition and include validated scale items for ground truth labeling. |
| 2 | Data Collection and Labelling via Self-Reports | Generate labelled training data using self-reported perceptions | Recruit respondents (e.g., Prolific, MTurk); collect their responses using both open-ended text and self-reported responses using validated scales | There is no fixed "N" but larger datasets (e.g., N > 1,000) improve robustness and class balance; mitigate imbalance using oversampling of the of minority class or SMOTE (Chawla et al., 2002; Khan et al., 2025) |
| 3 | Model Selection and Prompt Design | Select a base LLM and design input-output structure | Choose open-source LLM (e.g., LLaMA, Mistral) with adequate parameter size and inference cost; design classification prompts | 3B–7B parameter models are preferred for efficiency; Select the base LLM considering multilingual capability, training cost, tokenizer coverage |
| 4 | Supervised Fine-Tuning | Train the LLM on labelled data using parameter-efficient tuning | Apply Low-Rank Adaptation (LoRA); use AdamW optimizer; split dataset into training (60–65%), validation (15–20%), and test (20%) | Tune r (rank) = 8–16; learning rate to $\eta = 2 \times 10^{-4}$ (Loshchilov & Hutter, 2019); set max sequence length to 2048 tokens; use early stopping if validation loss does not improve after 3–5 evaluations |
| 5 | Out-of-Sample Performance Evaluation | Evaluate LLM performance on unseen labelled data | Measure macro-F1, weighted-F1; benchmark against existing approaches (e.g., LIWC, BERT, GPT) | F1 > 0.70 = acceptable; use scikit-learn or Hugging Face Eval libraries; report both macro and weighted F1 for imbalanced classes |
| 6 | Generalizability Testing (Domain Shift) | Test model robustness across relevant domains | Sample public data (e.g., Twitter, Amazon); annotate random subset using 2+ coders; resolve disagreements | Assess intercoder agreement; the F1 may increase in public data due to longer-form texts and the general-purpose capabilities of LLMs (Khan et al., 2025) |
| 7 | Nomological and Behavioral Validation | Validate the construct's role within a theoretical model | Use LLM-derived constructs as variables to extend scale-based research on limited samples | To assess nomological validity, test whether LLM-derived constructs predict outcomes, mediate theoretical pathways, or correlate with established constructs (e.g., ratings, purchases, satisfaction) |

*Note: Study 1 uses steps 1-5, Study 2 uses step 6, and Study 3 uses step 7.

*4.1 Data Collection and Survey Design*

Traditional methods for training LLMs to detect consumer emotions and sentiment rely on third-party text annotated by human coders, which creates several challenges. First, consumers who review a product or service generally describe only one or two perceptions, so limited informational content is available. Second, the coders' interpretations might not align with the author's original intent. Third, substantial coder disagreements demand extensive resolution efforts. For instance, Khan et al. (2025) report that coding 11,800 software reviews for eight emotions took seven hours per coder, with a 42.1% disagreement rate that required further reconciliation.

To overcome these limitations, we conducted a large-scale survey on Prolific, a platform known for high-quality data collection. We recruited 2,800 respondents from the United States, Canada, United Kingdom, and Australia, whom we compensated at a rate of £9 per hour for a 15-minute survey. Data quality checks led us to exclude 277 responses due to failed attention checks, time-outs, or incomplete answers, resulting in a final sample of 2,523 valid responses ($M_{age}$ = 38.6 years, $SD_{age}$ = 12.96; 42.0% men, 56.6% women, 1.4% preferred not to disclose gender; see Web Appendix B for further details of the respondent profile.). These respondents had to write about marketing-relevant themes in relation to a product or service they had experienced in the past month, then evaluate each theme according to an established scale. To ensure respondents had prior experience writing about product or service perceptions, we prescreened them to have active memberships on major review platforms such as Amazon, TripAdvisor, or Yelp. Respondents were asked to write about a product or service experience in one of nine categories: retail, clothing/footwear, furnishings, restaurant, health, transport, communication, recreation, or insurance (Table B3 indicates the number of respondents who selected each category).

Because the proposed LLM needs to detect multiple marketing-relevant themes, we structured the survey carefully, to capture respondents' perceived emotions, trust, commitment, recommendation likelihood, and sentiments toward marketing-relevant aspects (price, product/service, place, promotion, or 4Ps) of the product or service they chose to review. Respondents began by writing open-ended responses (of at least 15 words), elicited by the following question: "For the product/service you chose, please describe the emotions you felt when you were using the product/service." A new webpage then displayed their response, alongside a list of 16 consumption-relevant emotions from Richins (1997) (Table B4 contains the complete list of emotions). Respondents were then asked: "Which of the following emotions did you feel when you were using the product/service (you can choose multiple emotions)?" After describing and revealing their emotional reactions to the product or service, respondents completed similar description and evaluation exercises to reveal their perceptions of trust, commitment, recommendation likelihood, and sentiment toward marketing-relevant aspects (4Ps). Table 2 contains the instructions for all the emotion and evaluation themes.

Along with requesting open-ended text, we provided respondents with a measurement scale for each evaluation theme, each of which contained three items in random order; an exception is recommendation likelihood, for which we relied on Reichheld's (2003) single-item, 11-point scale (0 = "not at all likely to recommend," 10 = "extremely likely to recommend"). Most items used seven-point Likert scales (1 = "strongly disagree," 7 = "strongly agree). The complete list of measurement items in Table 3 displays their high Cronbach's alpha values, which confirm that the scales are reliable. As Figure 2 reveals, the combination of open-ended text input with structured scale responses ensures that respondents can refer to their own written text when they complete the scale items. Furthermore, the survey dynamically adjusts the measurement items depending on whether each respondent has chosen to review a product or service. For products, it includes all the emotions, recommendation likelihood, and aspect-based sentiment; for services, it also features trust and commitment.

**Table 2.** Written Text Generation Instructions for Each Perception in the Survey

| Perception | Text Generation Instructions |
|---|---|
| Consumer emotions | For the product/service you chose, please describe the emotions you felt when you were using the product/service.* |
| Trust | For your chosen service, please write about the degree that you trust the service and view it as being altruistic, reliable, honest, and competent.? In other words, do you trust the service or not? |
| Commitment | For your chosen service, please write about whether you believe your relationship with the product is so valuable that you are committed to this service for a long time? |
| Recommendation | For your chosen product/service, please write about whether you would recommend other customers to purchase/visit/use/get the particular product/service. |
| Consumer sentiment: price | Overall what are your thoughts about the price you were charged for the product/service? |
| Consumer sentiment: product | Overall what are your thoughts about the product/service quality and performance? |
| Consumer sentiment: place | Overall, what are your thoughts about the space or place where you obtained the product or received the service? |
| Consumer sentiment: communication | Overall, what are your thoughts about the communication between you and the firm and/or its employees that provided you with the product/service? |

*For the open-ended written text prompts, each instruction concluded with a parenthetic statement: "(minimum 15 words, please do not spend more than 2 minutes on this task)."

**Figure 2.** Example Survey Display: Written Text and Scale Items for Trust

For the service you chose, **please write about the degree that you trust Chipotle and view it as being altruistic, reliable, honest, and competent.** In other words, do you trust Chipotle or not? (minimum 15 words about 100 characters, please do not spend more than 2 minutes on this task).

> Overall, I trust Chipotle to do right by customers and to honor the commitments they make in their advertising. They use locally-sourced ingredients and actually pay their workers a living wage, which I appreciate.

--- Page Break ---

Listed below are statements pertaining to your experience with Chipotle. For each statement, please indicate how strongly you agree or disagree by selecting the appropriate box based on the review you wrote:

Overall, I trust Chipotle to do right by customers and to honor the commitments they make in their advertising. They use locally-sourced ingredients and actually pay their workers a living wage, which I appreciate.

| | Strongly disagree | Disagree | Somewhat disagree | Neither agree nor disagree | Somewhat agree | Agree | Strongly agree |
|---|---|---|---|---|---|---|---|
| Chipotle cannot be trusted at times. | ○ | ○ | ● | ○ | ○ | ○ | ○ |
| Chipotle can be counted on to do what is right. | ○ | ○ | ○ | ○ | ● | ○ | ○ |
| Chipotle has high integrity. | ○ | ○ | ○ | ○ | ● | ○ | ○ |

**Table 3.** Scale Items for Each Perception in the Survey

| Perception | Question and Scale Items | Reference |
|---|---|---|
| Consumer emotions* | Which of the following emotions did you feel when you were using the product/service (You can choose multiple emotions; example response options "Joy: Happy, Pleased, Joyful.") | Richins (1997) |
| Trust $\alpha = .88$ | **Question:** For each statement below, please indicate how strongly you agree or disagree by selecting the appropriate box: <br> The product/service cannot be trusted at times. <br> The product/service can be counted on to do what is right. <br> The product/service has high integrity. | Morgan and Hunt (1994) |
| Commitment $\alpha = .93$ | **Question:** For each statement below, please indicate how strongly you agree or disagree by selecting the appropriate box: <br> I am very committed to the product/service. <br> I intend to use the product/service indefinitely. <br> My relationship with the product/service provider deserves maximum effort to maintain. | Morgan and Hunt (1994) |
| Recommendation* | **Question:** How likely would you recommend this product/service to a friend? <br> (0 = "not at all likely" to 10 = "extremely likely") | Reichheld (2003) |
| Consumer sentiment: price $\alpha = .94$ | The product/service I bought was overpriced. <br> The price for the product/service was fair. <br> In general, I am satisfied with the price I paid. | Gaski and Etzel (1986) |
| Consumer sentiment: product $\alpha = .86$ | The quality of the product/service is as good as can be expected. <br> I am satisfied with the product/service. <br> The business that made the product/provided the service doesn't care enough about how well they perform. | Gaski and Etzel (1986) |
| Consumer sentiment: place $\alpha = .85$ | I am satisfied with the place where I bought/used the product/service or how it was delivered. <br> I have problems with or complaints about the place where I bought/used the product/service or how it was delivered. <br> Because of the place where I bought/used the product/service or how it was delivered most of my experience was pleasant. | Gaski and Etzel (1986) |
| Consumer sentiment for communication $\alpha = .81$ | The communication was very annoying. <br> I enjoyed the communication. <br> If the communication was eliminated, I would have been better off. | Gaski and Etzel (1986) |

Notes: The full list of emotions and their definitions appears in Table B4.
*Cronbach alpha values are not calculated for consumer emotions and recommendation because they use single-item scales.

## 4.2 Model Development

In the model development process, we fine-tune LX using the survey data set and prepare it for the multitask classification across consumer emotions and evaluations. Most LLMs are pretrained on extensive factual corpora, such as Wikipedia, to predict subsequent tokens in a text sequence based on preceding tokens (Alizadeh et al. 2025; Arora,

Chakraborty, and Nishimura 2025). Such models like Llama 3.2 and GPT 4-Turbo exhibit strong general-purpose capabilities across various tasks (Alizadeh et al. 2025; Khan et al. 2025) but fine-tuning effectively enhances their performance for specific applications. Fine-tuning involves three steps: selecting a pretrained baseline LLM, prompt engineering to develop effective instructions to guide the LLM training, and parameter-efficient fine-tuning to improve prediction accuracy and computational efficiency by refining the model parameters.

*4.2.1 Baseline LLM*

Since being introduced by Touvron et al. (2023), Meta AI's open-source Llama series has evolved, so we utilize the latest version, Llama-3.2-3B-Instruct, as a base model. Llama 3.2 offers several key benefits: it is cost-free, sufficiently open-source to allow users to retain their private training data when undertaking fine-tuning, and multilingual, such that it supports task deployment in more than 30 languages (Kili Technology 2025). Although an 8-billion-parameter version exists, we opt for the 3-billion-parameter model, due to its advantageous memory usage, training flexibility, and accurate inferences for practical development.

*4.2.2 Prompt engineering*

We design the prompts explicitly to detect constructs that have been validated in marketing theory, such as Richins's (1997) emotion set and Morgan and Hunt's (1994) trust scale. Therefore, the model outputs will correspond to theoretically meaningful perceptions. We input consumer text, a prompt, and three tasks: emotion detection, evaluation polarity detection, and aspect-based sentiment analysis, as detailed in Table 4.

**Table 4.** Tasks, Instructions, Perceptions, and Possible Outcomes from the LLM

| Task | Instruction (prompt) | Perception | Possible output |
|------|----------------------|------------|-----------------|
| Emotion detection | Detect emotion $\langle E_1 \cdots E_{16} \rangle$ present or not in the given text | Joy | $E_1 \cdots E_{16}$ Present |
| | | Anger | |
| | | $\vdots$ | $E_1 \cdots E_{16}$ Not Present |
| | | Surprise | |
| Polarity detection | Determine the polarity for perception theme $\langle TH_1 \cdots TH_3 \rangle$ in the given text. Choose from $\langle TH_j \rangle$ positive, negative, or neutral | Trust | $TH_1 \cdots TH_3$ Positive |
| | | Commitment | $TH_1 \cdots TH_3$ Negative |
| | | Recommendation | Neutral/No Mention |
| Sentiment analysis | Determine the sentiment for marketing aspect $\langle P_j \rangle$ in the given text. Choose from $\langle P_j \rangle$ positive, negative, or neutral | Price | $P_1 \cdots P_4$ Positive |
| | | Place | |
| | | Promotion† | $P_1 \cdots P_4$ Negative |
| | | Product | Neutral/No Mention |

†We use the term "Promotion" here to reflect the common 4Ps marketing terminology, but in the survey, we used the term "Communication."

First, an initial task entailed detecting any of 16 emotions within the provided text. For example, if the assigned task was to detect the emotion of "joy," the straightforward prompt would read, "detect joy present or not in the text." After processing the assigned text (e.g., "So impressed! The quality is outstanding, and it feels like it was made just for me. I trust this brand and will definitely buy from them again"), the LLM selects from two possible outcomes: "joy present" or "joy not present." In the survey, respondents could list up to 16 emotions they felt during their product or service experience, and if they had indicated feeling "joy," the LLM would be trained to predict this emotion accurately.

Second, unlike emotion detection, which is a binary task, detecting the polarity of marketing-relevant evaluations (e.g., trust, commitment, recommendation likelihood) requires survey data spanning several scale items, so that it is possible to determine whether a respondent's evaluation is positive, negative, or neutral. In the example in Figure 2, after reverse-coding responses to the first item, the values for the three trust items get averaged to obtain the score for each respondent. Average scores less than 4 are negative, those equal to 4 are neutral, and if they are greater than 4, it indicates a positive evaluation.[2] However, the recommendation construct requires a different approach, so we adopt Reichheld's (2003) classification (and labels): Scores of 6 or less are negative (detractors), 7 and 8 are neutral (passives), and 9 or higher are positive (promoters). To map the LX outputs to these validated constructs, we apply standardized classification rules, derived from established marketing scales.

Third, for the aspect-based sentiment analysis, we seek to categorize the survey responses as they relate to each aspect of the marketing 4Ps. For example, Table 3 shows that price-related sentiment, is measured on a 7-point Likert scale with three items from Gaski and Etzel's (1986) price perceptions construct: (1) "The product/service I bought was overpriced," (2) "The price for the product/service was fair," and (3) "In general, I am satisfied with the price I paid." After reverse coding the first item, the LLM averages the scores and categorizes them: –1 for averages less than 4, 0 for exactly 4, and 1 for greater than 4. Summing these categorizations across all 4Ps produces an overall sentiment score, such that it is negative for sums less than 0, neutral for exactly 0, and positive for greater than 0. With these task definitions and labels, we can fine-tune the LX using supervised learning techniques, optimized for multitask performance.

*4.2.3 Training procedure*

The supervised fine-tuning of the LLM entails an instruction-tuning process across multiple tasks, which enables the LLM to produce task-specific responses. Formally, the instruction data set is defined as $\mathcal{D} = \{(x, I^{(t)}, y^{(t)}), t \in T\}$, where $t$ denotes a specific task within the task set $T$, $I^{(t)}$ refers to the instructions for task $t$, and $x$ and $y^{(t)}$ are, respectively, the input and target output for task $t$. The objective is to generate a correct response $(y^{(t)})$ given the prompt and input $(I^{(t)}, x)$. A machine learning model can be viewed like a function $f(\cdot)$ that predicts an outcome $y$ given the prompt $(I^{(t)}, x)$ with a set of parameters (or weights) denoted by $W$. Then the prediction model can be written as

$$y = f_W(I^{(t)}, x). \tag{1}$$

---

[2] In the survey, all respondents evaluated trust, commitment, recommendation likelihood, and the 4Ps. However, in the Amazon and Yelp reviews in Study 2, not all of these themes were present in each text, so those data sets include an additional "No Mention" output.

A pretrained LLM with parameters $W$ predicts $y$ for the input data. Any differences between the model prediction and the ground truth can be gauged with a loss function, such as the cross-entropy function (Feng et al. 2024; Sengupta et al. 2024):

$$\mathcal{L} = -\sum_{i=1}^{K} \log\left(\Pr\left(y_i^{(t)} | y_{<i}^{(t)}, (I^{(t)}, x)\right)\right), \tag{2}$$

where $y_i^{(t)}$ is the $i$th token in the output sequence, $y_{<i}^{(t)}$ is all the tokens preceding $y_i^{(t)}$, and $K$ is the total number of tokens in the output sequence $y^{(t)}$. For the example emotion detection task involving joy, the ground truth output sequence is "joy present," with two tokens, so $K = 2$. The probabilities in this equation are determined by the LLM and are conditional probabilities for the next token in the sequence, given the input data, instructions, and output so far. The conditional probabilities also depend on the LLM parameters $W$, so fine-tuning involves adjusting $W$ to minimize the loss function.[3] To achieve loss minimization we use stochastic gradient descent, which iteratively updates the parameters in the direction of the negative gradient:

$$W \leftarrow W - \eta \Delta W, \tag{3}$$

where $\Delta W = \frac{\partial \mathcal{L}}{\partial W}$ is the vector gradient of model parameters, and $\eta$ is the learning rate that controls the step size for each update. Web Appendix C contains an illustration of the reduction of loss function after adjusting the parameter vector $W$.

*4.2.4 Parameter-efficient fine tuning*

---

[3] The loss function in Equation 2 for each individual text is summed across all texts in the training data set.

Fine-tuning an LLM with many parameters is computationally expensive. To mitigate these costs, we employ low-rank adaptation (LoRA) for parameter-efficient fine-tuning (Hu et al. 2021; Mangrulkar et al. 2022; Ye, Yoganarasimhan, and Zheng 2025). LoRA freezes the pretrained weights $W$ and decomposes the weight updates $\Delta W$ in Equation 3 into two smaller, low-rank matrices, $A$ and $B$. Thus, $\Delta W = B \times A$, where $W \in \mathbb{R}^{m \times n}, A \in \mathbb{R}^{r \times n}$, and $B \in \mathbb{R}^{m \times r}$. The rank $r$ is typically low, with $r \ll \min(n, m)$, where $n$ and $m$ indicate the input and output dimensions of $W$, respectively (Sengupta et al. 2024). Based on a grid search, we determine $r$ to be equal to 16, and we identify a scale factor $\alpha$ = 32 for $\Delta W$. According to Hu et al. (2021), applying LoRA to GPT-3 reduces the number of trainable parameters by a factor of 10,000 and cuts memory requirements by two-thirds.

During training, we set the maximum input length to 2,048 tokens, though none of the survey texts approached this length. We optimize the model using the AdamW optimizer with a linear learning rate scheduler; we set the learning rate in Equation 3 to $\eta = 2 \times 10^{-4}$ as per Loshchilov and Hutter (2019). The split of the survey data into training, validation, and test sets imposes proportions of .64, .16, and .20, respectively.[4] Training is conducted for 3000 iterations of Equation (3), with evaluations on the validation set every 100 iterations. This design uniquely avoids inferential errors from third-party labeling by training directly on perceptions as reported by consumers themselves.

*4.2.5 Deploying Llama3.2-3B-Instruct*

The deployment of Llama 3.2-3B-Instruct involves constructing prompts using the same format as our instruction data set, with an added sentence that lists the output categories of the target perception theme as options. This guidance leads the model to generate an option letter instead of a full sentence, for simplified interpretation and processing. For the previously established joy example, the instruction would read "Detect emotion (Joy) present or not in the given text. Provide your answer by choosing the option letter: A. Joy is present in the text, or B. Joy is not present in the text." Then the potential outputs are A or B.

---

[4] In the survey, respondents mention emotions with varying frequencies; for example, joy was noted by 939 people, whereas fear was mentioned by 151. Therefore, this data partitioning occurs separately for each emotion. We apply similar stratifications to the polarity detection and sentiment analysis tasks.

*4.2.6 Handling data imbalance*

In our survey data, 939 of 2,523 respondents reported experiencing joy, but only 39 (1.5%) reported loneliness. Such imbalance can lead models to favor a majority class, resulting in inaccurate predictions for minority classes. We therefore apply oversampling and undersampling techniques, as recommended by Chawla et al. (2002) and Khan et al. (2025). Specifically, we match 39 instances of "loneliness present" with an equal number of "loneliness not present" cases to balance the training data.[5]

*4.3 Baseline and Benchmark Models*

To benchmark the performance of our fine-tuned LLM, which we refer to as the LX, we compare it against widely used transformers and proprietary LLMs, capable of similar tasks. Both the BART[6] and RoBERTa[7] transformer models can undertake the tasks listed in Table 4, the recently-released pretrained LLMs Llama 3.2-8B-Instruct, GPT 3.5-turbo-instruct, GPT 4-Turbo and Deep Seek-V3 are all capable of detecting emotions and measuring consumer evaluations, such as sentiment. Web Appendix E lists each of these models, and details their number of parameters and associated costs, if they apply.

*4.4 Performance Evaluation*

The comparative evaluation, involving 20 marketing-relevant perception tasks, relies on an out-of-sample data set of open-ended survey responses labeled with both binary emotion classifications and categorized evaluations (negative, neutral, or positive). Because multiple emotions can appear in a single response, the LX model detects each emotion separately. Model performance is evaluated using the F1 score, the harmonic mean of precision and recall (Berger et al. 2020), being standard in classification tasks. To capture overall performance, we report both (i) Macro F1 (the unweighted average across 20 perception classification tasks) and (ii) Weighted F1 (which accounts for the relative size of each task in the test data).

---

[5] To address the underrepresentation of neutral evaluations in performance themes and sentiment aspects, and ensure a balanced data set across positive, negative, and neutral categories, we also supplement the survey training data with randomly selected, neutral-category reviews from Amazon and Yelp. The emotions data set does not require supplementation though, because it contains sufficient survey responses. Web Appendix D details the quantity of texts for each outcome in the training and test data sets.
[6] See https://huggingface.co/facebook/bart-large-mnli
[7] See https://huggingface.co/FacebookAI/roberta-large-mnli

Table 5 compares the LX model's classification accuracy, using the F1 score, across emotion and evaluation polarity tasks. The LX model achieves the best or equal-best performance in 13 of the 16 emotion categories. For polarity detection among negative, neutral, and positive categories, LX consistently yields the highest F1 scores, outperforming larger models including GPT variants and Deep Seek.[8] Both Macro and Weighted F1 scores confirm LX's greater accuracy with the test data. Compared with the pretrained but non–fine-tuned Llama 3.2-3B, LX exhibits consistent improvement, underscoring the effectiveness of the fine-tuning process.

As a further benchmarking exercise, we also apply the LIWC (Pennebaker et al. 2015) to detect anxiety (worry), anger, and sadness, the only emotions available in the LIWC dictionary. The F1 scores of .62, .74, and .65, respectively, are substantially lower than those achieved by LX. Similarly, VADER sentiment analysis (Hutton and Gilbert 2014) produces an overall F1 score of .51, similar to the transformer models but well short of LX's .88 score.

---

[8] For the three polarity outcome classes, we calculate F1 for each class separately by treating all other classes as one class, thus converting the problem into a binary classification. We report the average F1 values in Table 5 (see https://scikit-learn.org/1.1/modules/generated/sklearn.metrics.classification_report.html).

**Table 5.** LLM Classification Accuracy for Emotions and Evaluations (Survey Data)

| Perception | Test Size | BART | RoBERTa | Llama3.2-3B | Llama3.2-8B | GPT-3.5-Turbo | GPT-4-Turbo | DeepSeek-V3 | LX[†] |
|---|---|---|---|---|---|---|---|---|---|
| **Consumer Emotions** | | | | | | | | | |
| Joy | 374 | 0.51 | 0.68 | 0.83 | 0.85 | **0.86** | 0.84 | 0.84 | **0.86** |
| Worry | 157 | 0.6 | 0.66 | 0.69 | **0.76** | 0.71 | 0.73 | **0.76** | **0.76** |
| Contentment | 392 | 0.77 | 0.73 | 0.8 | 0.83 | 0.82 | 0.83 | **0.84** | 0.82 |
| Envy | 12 | 0.38 | 0.49 | 0.33 | 0.33 | 0.62 | 0.33 | 0.33 | **0.73** |
| Anger | 401 | 0.46 | 0.55 | 0.77 | 0.84 | 0.9 | 0.86 | 0.84 | **0.91** |
| Surprise | 106 | 0.5 | 0.39 | 0.41 | 0.43 | 0.56 | 0.51 | 0.44 | **0.67** |
| Fear | 60 | 0.48 | 0.68 | 0.65 | 0.69 | 0.71 | 0.71 | 0.69 | **0.92** |
| Shame | 63 | 0.39 | 0.49 | 0.5 | 0.58 | **0.73** | 0.55 | 0.45 | 0.69 |
| Loneliness | 15 | 0.55 | 0.7 | 0.55 | 0.64 | 0.44 | 0.55 | 0.55 | **0.72** |
| Excitement | 251 | 0.66 | 0.61 | 0.66 | 0.73 | 0.7 | 0.74 | 0.68 | **0.79** |
| Peacefulness | 260 | 0.74 | 0.76 | 0.71 | 0.72 | 0.57 | 0.7 | 0.65 | **0.81** |
| Love | 120 | 0.57 | 0.6 | 0.67 | 0.76 | 0.73 | 0.71 | 0.74 | **0.83** |
| Romantic love | 21 | 0.28 | 0.31 | 0.32 | 0.42 | 0.42 | 0.42 | 0.5 | **0.86** |
| Discontent | 360 | 0.64 | 0.75 | 0.82 | 0.86 | 0.84 | 0.85 | **0.87** | **0.87** |
| Sadness | 120 | 0.74 | 0.76 | 0.66 | 0.75 | 0.76 | **0.81** | 0.75 | 0.79 |
| Optimism | 214 | 0.62 | 0.67 | 0.69 | 0.7 | 0.68 | 0.67 | 0.72 | **0.74** |
| **Consumer Evaluations** | | | | | | | | | |
| Trust | 403 | 0.49 | 0.49 | 0.49 | 0.72 | 0.73 | 0.81 | 0.75 | **0.9** |
| Commitment | 397 | 0.47 | 0.5 | 0.55 | 0.65 | 0.73 | 0.73 | 0.73 | **0.9** |
| Recommendation | 409 | 0.49 | 0.5 | 0.51 | 0.56 | 0.54 | 0.58 | 0.53 | **0.74** |
| Sentiment[†] | 1,611 | 0.46 | 0.47 | 0.6 | 0.66 | 0.69 | 0.71 | 0.72 | **0.88** |
| **Overall Accuracy** | | | | | | | | | |
| Macro F1 | 5746 | 0.54 | 0.59 | 0.61 | 0.67 | 0.69 | 0.68 | 0.67 | **0.81** |
| Weighted F1 | 5746 | 0.54 | 0.57 | 0.65 | 0.71 | 0.72 | 0.74 | 0.73 | **0.84** |
| Cost[††] | - | 0 | 0 | 0 | 0 | $1.10 | $7.99 | $0.17 | 0 |

Notes: Values in bold indicate the best predictive model, based on the F1 statistic.
[†] The F1 score for sentiment is the macro F1 value across all four marketing aspects.
[††] The number of input and output tokens are, respectively, 676,387 and 40,804.

*4.5 Discussion*

Study 1 results indicate that LX accurately detects self-reported consumer emotions and evaluations in open-ended text. It outperforms both traditional and cutting-edge models, with an overall F1 measure that is more than 10 percentage points higher than the next-best models. Furthermore, it does not incur costs, unlike the GPT and Deep Seek models. In our example, the test data set includes only 2,523 responses, averaging 268 tokens each (around 200 words). Scaling this process to a more typical scenario, such as one with 500,000 reviews of similar length, would raise GPT 4-Turbo's cost to approximately $1,585. Having established this promising foundation, we next evaluate LX's generalizability to third-party–annotated data.

## 5. Study 2: Benchmarking LX Using Publicly Available Review Data

To evaluate the generalizability of the LX model, we apply it to actual, consumer-authored reviews from Amazon and Yelp. Building on Study 1's demonstration of LX's ability to detect validated emotions and evaluations from consumers' self-reported survey text, Study 2 tests for LX's accuracy when applied to third-party review data, which is critical for assessing the external and construct validity of the model. We seek to confirm its ability to identify marketing-relevant consumer perceptions, such as discontent, trust, and recommendation likelihood, from observed (rather than surveyed), unstructured, unsolicited consumer input.

*5.1 Data Sources and Annotation Protocol*

This study leverages two publicly available review data sets that have been employed in prior marketing and information systems research: Amazon product reviews (Ni, Li, and McAuley 2019; https://nijianmo.github.io/amazon/) and Yelp restaurant reviews (www.yelp.com/dataset). From the Amazon data set, we randomly sampled 8,000 sentences for emotions, 8,000 for sentiment, and 2,000 for trust, commitment, and recommendation likelihood. We then excluded non-English sentences and statements unrelated to the product or service. This left 7,450 sentences coded for the presence or absence of each emotion, 7,306 sentences coded for sentiment (positive or negative), and 1,824 sentences coded for positive or negative mentions of trust, commitment, and recommendation. Because we did not apply any filters related to the product category, review length, or ratings, these samples support broad, representative evaluations of LX's performance across text types and tones.

To establish the labeled training and test sets, we implemented a multistep human annotation protocol. With a custom-built interface in Prodigy3, annotators coded sentences for the emotion detection, polarity detection for evaluations, and aspect-based sentiment detection for the four Ps (price, product, place, and promotion) tasks. For each sentence, coders first determined if a specific perception was present; if so, they assigned it a polarity label (negative = –1, neutral/no polarity = 0, positive = 1). Each sentence was independently annotated by two coders, who had undergone extensive training and calibration. The coders completed multiple training rounds (100 sentences per coder per task), involving two reconciliation rounds, after 250 sentences and then after a further 250 sentences. They annotated the remainder of the data independently. This protocol ensured consistent, theoretically meaningful construct labels, aligned with marketing literature.

*5.2 Intercoder Agreement*

To gauge intercoder agreement (ICA), we determined the proportion of cases in which both annotators fully agreed on both the specific emotion or evaluation being present (yes = 1, no = 0) and, for the evaluations, concurred about their polarity. As Table 6 reveals, the ICA values were very high for most emotion and evaluation perceptions; they were lower but still acceptable in relation to more complex or composite constructs, such as contentment, discontent, and sentiment. To ensure the strong quality of the data in the final training and test review data sets, we only include instances for which both coders fully agreed. For example, 94% of the 7,450 sentences labeled for "joy" showed full agreement, and for 514 sentences they coded yes (1) for the presence of joy across all the Amazon and Yelp reviews. We used half of these agreed coding observations for further training and the other half for testing. Furthermore, we balance the training and test data for the presence or absence of each emotion. For instance, Table 7 shows there are 458 sentences used when testing for "joy" in the Amazon test data set. In this case 229 have joy present while the other 229 do not. We train LX, according to the same procedure we described for the survey data, except that in this case, the ground truth is established by human coding.

Table 6. Intercoder Agreement for Online Review Data

| Consumer Perception | Proportion of Sentences for Which Coders Agree |
|---|---|
| ***Consumer Emotions*** | |
| Joy | .94 |
| Worry | .99 |
| Contentment | .80 |
| Envy | 1.00 |
| Anger | .97 |
| Surprise | .99 |
| Fear | .99 |
| Shame | 1.00 |
| Loneliness | 1.00 |
| Excitement | .99 |
| Peacefulness | .99 |
| Love | .99 |
| Romantic love | .99 |
| Discontent | .87 |
| Sadness | .98 |
| Optimism | .98 |
| ***Consumer Evaluations*** | |
| Trust | .98 |
| Commitment | .97 |
| Recommendation | .98 |
| Sentiment | .86 |

*5.3 Results for Amazon and Yelp Reviews*

For the LX evaluations across both data sets, we use the transformer and LLM baselines introduced in Study 1. The results for the Amazon test data set in Table 7 confirm that LX achieves the highest or equivalent F1 scores in nearly every perception category, both emotional (e.g., joy, anger, sadness) and evaluative (e.g., trust, recommendation). Thus, we can confirm LX's ability to detect a broad range of marketing-relevant consumer perceptions. The strong results in each category translate into superior overall accuracy, such that LX attains a macro F1 of .97 and a weighted F1 of .95, substantially better than all benchmark models.[9]

---

[9] Interestingly, the F1 statistics for all the LLMs are higher for the Amazon and Yelp data than for our survey data. This is because LLMs are typically pretrained on such publicly available consumer reviews.

**Table 7.** Perception-Level F1 Scores and Accuracy, Amazon Review Test Data

| Perception | Test Size | BART | RoBERTa | Llama3.2-3B | Llama3.2-8B | GPT-3.5-Turbo | GPT-4-Turbo | DeepSeek-V3 | LX |
|---|---|---|---|---|---|---|---|---|---|
| **Consumer Emotions** | | | | | | | | | |
| Joy | 458 | .42 | .57 | .81 | .83 | .8 | .76 | .77 | **.96** |
| Worry | 68 | .53 | .58 | .93 | .87 | .8 | .91 | .93 | **1** |
| Contentment | 946 | .56 | .53 | .62 | .72 | .73 | .76 | .76 | **.9** |
| Envy | 46 | .5 | .5 | .98 | .98 | .98 | .98 | 1 | **1** |
| Anger | 156 | .4 | .5 | .88 | .91 | .9 | .94 | .96 | **.97** |
| Surprise | 320 | .46 | .38 | .97 | .94 | .81 | .92 | **.99** | .99 |
| Fear | 86 | .41 | .59 | .97 | **1** | .99 | 1 | 1 | 1 |
| Shame | 132 | .41 | .39 | .92 | .97 | .86 | .95 | .97 | **1** |
| Loneliness | 0 | - | - | - | - | - | - | - | - |
| Excitement | 156 | .53 | .58 | **.97** | .95 | .9 | .88 | .92 | **.97** |
| Peacefulness | 32 | .51 | .65 | .94 | .91 | **1** | .97 | .97 | .97 |
| Love | 516 | .54 | .55 | .95 | .94 | .91 | .94 | .9 | **.98** |
| Romantic love | 136 | .53 | .46 | .86 | .84 | .69 | .68 | .52 | **.99** |
| Discontent | 296 | .4 | .5 | .86 | .89 | .84 | .87 | .87 | **.9** |
| Sadness | 174 | .57 | .58 | .83 | .89 | .89 | .91 | .95 | **.99** |
| Optimism | 150 | .52 | .64 | .78 | .77 | .76 | .76 | .75 | **.99** |
| **Consumer Evaluations** | | | | | | | | | |
| Trust | 66 | .52 | .5 | .47 | .75 | .75 | .84 | .74 | **.95** |
| Commitment | 84 | .53 | .53 | .54 | .75 | .79 | .47 | .87 | **.98** |
| Recommendation | 132 | .53 | .54 | .53 | .66 | .75 | .88 | .64 | **.99** |
| Sentiment[†] | 177 | .6 | .53 | .62 | .71 | .61 | .67 | .64 | **.82** |
| **Overall Accuracy** | | | | | | | | | |
| Macro F1 | 4131 | .5 | .53 | .81 | .86 | .83 | .85 | .85 | **.97** |
| Weighted F1 | 4131 | .5 | .53 | .79 | .84 | .81 | .83 | .83 | **.95** |

Notes: Values in bold indicate the best predictive model, based on the F1 statistic.
[†] The F1 score for sentiment is the macro F1 value across the four marketing aspects.

The results for the Yelp data set in Table 8 indicate that, even for shorter, more informal text (as is common on this platform), LX demonstrates better classification accuracy for most categories. Its macro F1 value is .96, and its weighted F1 value is .89. It performs especially well for high-frequency emotions (e.g., joy, anger, contentment) and core evaluations (e.g., trust, commitment), but it also offers good results for rare and ambiguous constructs, such as peacefulness, romantic love, and envy, that generally are subject to data sparsity. Such consistent performance across both review platforms supports the generalizability and robustness of LX.

**Table 8.** Perception-Level F1 scores and Overall Accuracy, Yelp Review Test Data

| Perception | Test Size | BART | RoBERTa | Llama3.2-3B | Llama3.2-8B | GPT-3.5-Turbo | GPT-4-Turbo | DeepSeek-V3 | LX |
|---|---|---|---|---|---|---|---|---|---|
| **Consumer Emotions** | | | | | | | | | |
| Joy | 56 | .41 | .67 | .78 | .84 | .89 | .89 | .89 | **.91** |
| Worry | 6 | .62 | .33 | .83 | .83 | .62 | .62 | 1 | **1** |
| Contentment | 124 | .75 | .66 | .76 | .85 | .82 | .87 | .87 | **.95** |
| Envy | 14 | .47 | .59 | .93 | **1** | **1** | **1** | 1 | **1** |
| Anger | 64 | .4 | .54 | .91 | .89 | .81 | .91 | .91 | **.94** |
| Surprise | 48 | .45 | .42 | **1** | .96 | .81 | .89 | 1 | .98 |
| Fear | 0 | - | - | - | - | - | - | - | - |
| Shame | 26 | .33 | .33 | .88 | .92 | .8 | **.96** | .92 | **.96** |
| Loneliness | 2 | .33 | .33 | **1** | **1** | **1** | **1** | 1 | **1** |
| Excitement | 20 | .44 | .44 | **.95** | .79 | .79 | .6 | .79 | **.95** |
| Peacefulness | 40 | .64 | .76 | .92 | .9 | **.97** | .95 | **.97** | .95 |
| Love | 26 | .55 | .55 | .76 | .88 | .76 | .84 | .8 | **1** |
| Romantic love | 10 | .52 | .33 | .9 | **1** | .9 | .9 | 1 | **1** |
| Discontent | 80 | .45 | .58 | .85 | .82 | .86 | .83 | .82 | **.91** |
| Sadness | 36 | .62 | .58 | .94 | .97 | .97 | .92 | .94 | **1** |
| Optimism | 16 | .56 | .65 | .87 | .73 | .73 | .73 | .81 | **1** |
| **Consumer Evaluations** | | | | | | | | | |
| Trust | 24 | .5 | .39 | .46 | .64 | .78 | .7 | .74 | **.96** |
| Commitment | 24 | .53 | .53 | .53 | .66 | .65 | .31 | .78 | **.96** |
| Recommendation | 84 | .53 | .59 | .54 | .68 | .71 | .9 | .69 | **.93** |
| Sentiment[†] | 558 | .54 | .52 | .62 | .71 | .57 | .67 | .51 | **.81** |
| **Overall Accuracy** | | | | | | | | | |
| Macro F1 | 1258 | .51 | .52 | .81 | .85 | .81 | .82 | .87 | **.96** |
| Weighted F1 | 1258 | .54 | .55 | .72 | .78 | .71 | .77 | .7 | **.89** |

Notes: Values in bold indicate the best predictive model, based on the F1 statistic.
[†] The F1 score or sentiment is the macro F1 value across the four marketing aspects.

## 5.4 Benchmark Comparisons

We also compare LX's performance with that achieved in recent studies that apply LLMs to emotion detection tasks. For example, Feng et al. (2024) apply Llama 3-8B and GPT-4 to human dialogue and seek insights into seven emotion categories; they report F1 scores ranging from .36 to .64. When Peng et al. (2023) attempt to evaluate five emotions, using multilingual dialogue and social media data sets, the macro F1 scores range from .29 to .95, with an average of .66. Because the macro F1 scores we obtain for LX (.97 on Amazon and .96 on Yelp) exceed these benchmarks by more than 30 percentage points, this comparison helps demonstrate the value of theory-aligned fine-tuning based on validated marketing constructs. The results also indicate that LX generalizes better to consumer-authored review content than do models trained exclusively on open-domain or conversational data.

*5.5 Discussion*

According to Study 2, LX generalizes effectively to third-party online review data and maintains high classification accuracy for both product and service domains. The results are robust, though we also note that performance for low-frequency constructs (e.g., envy, loneliness) remains modest, in part due to their limited representation in the training and test data. Because the annotations refer to the sentence level, Study 2 also might ignore some contextual information embedded across longer text passages (as was the case in Study 1). Therefore, in Study 3 we examine whether LX-detected consumer perceptions generalize across text sources and also if they can predict downstream marketing-relevant outcomes, including product ratings and purchase behavior, on the basis of real-world clickstream and transaction data.

## 6. Study 3: Linking LX-Derived Emotions to Product Ratings and Purchase Behavior

Studies 1 and 2 established LX's accuracy and generalizability for detecting validated consumer perceptions; Study 3 evaluates whether emotions extracted by LX from consumer-authored reviews can predict downstream key marketing outcomes with behavioral relevance. With a mediation model, we postulate that LX-detected emotions are linked to product ratings, which in turn are associated with the likelihood of purchase among subsequent consumers who read those reviews. This behavioral validation of LX also extends prior research on the role of customer reviews in shaping consumer purchase decisions (e.g., Liu, Lee, and Srinivasan 2019;

Vana and Lambrecht 2021; Varga and Albuquerque 2024). Linking text-extracted consumer perceptions to product evaluations and purchase behavior can contribute to broader theories of affective influence. To the best of our knowledge, this study represents the first attempt to demonstrate that fine-tuned LLM outputs, based on validated emotions, can explain real-world consumer behavior in transactional settings.

*6.1 Data*

To identify the relevance of various emotions in consumer decision-making, we leverage customer browsing and purchase data from Wharton Customer Analytics, which pertain to a large, U.K., online retailer.[10] We obtain individual-level clickstream and transaction information for two broad departments—home and garden (e.g., sheets, pillows) and technology (e.g., TVs, printers)—for February and March 2015. Thus, we can track consumer browsing journeys, which usually start with a category search that then gets refined to particular products. Each product page displays the price, average rating and volume of reviews, and a product description, but detailed review text is not automatically displayed. Customers must scroll down to the bottom of the product page to find a tab they can click to read the full text of reviews. Because of this design, we can verify whether each user clicked to read review text. Of the 30,879 products in this data set, not all of them were browsed during the two-month window, and nearly 20% of products have no reviews. We limit our assessment to 10,491 products for which reviews are available and read by at least one user.

Reviews appear in batches. The first batch contains 5 reviews (or less, if fewer than 5 reviews are available), and subsequent batches contain 20 reviews, displayed only if a user requests more.[11] We cannot identify which reviews within a batch actually get read, so we assume they are all read, which seems reasonable given that the vast majority of batches have just 5 reviews and the average reading time for a batch of 5 reviews is 25 seconds.[12] We used LX to extract perceptions for 167,132 product reviews that were read by at least one person in the data window. To reflect the varying numbers of reviews for each product, we create a vector of emotions ($Emotions_i$) that contains the average occurrence for each emotion, as extracted by LX, across all reviews for product $i$. For example, if LX determines that, among 20 reviews for a product, 5 include expressions of anger, then the average anger score is .25 across all these reviews.

---

[10] To obtain these data, we agreed that the name of the retailer and an independent third-party company that managed the reviews would not be disclosed. Previous studies (Liu, Lee, and Srinivasan 2019; Vana and Lambrecht 2021; Varga and Albuquerque 2024) have used these data to demonstrate causal links between reading a review and subsequent purchase; none of them examine the emotional content of the review text.
[11] In 85% of cases, review readers consider only the initial batch of 5 or fewer reviews and do not request any additional reviews.

*6.2 Model*

In line with the available data, we propose a conceptual model in which a user's product rating reflects the emotions expressed in their review, and, in turn, product ratings are linked to purchases for those reading a product's reviews. We use the following system of equations that are based on this conceptual model. The first equation, for average rating, has a vector containing the 16 emotions as the independent variable:

$$Average\ Rating_i = \gamma_0 + \gamma_1 Emotions_i + \epsilon_i^R, \qquad (4)$$

where $Average\ Rating_i$ is the average star rating (out of 5) across all reviews for product $i$, and $i = 1, 2, \ldots, 10{,}491$ products.

The second equation relates a product's average rating to purchase behavior. We define $Purchase_i$ as the proportion of times that a visitor to the retailer's website reads a review for product $i$ and then purchases the product within their browsing session. For instance, if a particular product review is read 100 times (by different users) and on three occasions the user subsequently purchases the product within their session, then $Purchase_i = .03$. For this equation, we also account for critical control variables that likely inform the association between online reviews and purchases, namely, product price ($Price_i$); review volume ($Volume_i$); product popularity, measured as the number of times reviews for the product have been read ($Views_i$); and the average number of words in all reviews of that product ($Length_i$). Noting the presence of large values and skewness, we take the logs of these four control variables. Formally,

$$Purchase_i = \beta_0 + \beta_1 Average\ Rating_i + \beta_2 Price_i + \beta_3 Volume_i + \beta_4 Views_i + \beta_5 Length_i + \beta_6 Emotions_i + \epsilon_i^P. \tag{5}$$

---

[12] We set a minimum time to read a batch of reviews to 5 seconds, which ensures the batch of reviews is loaded and served to the customer and also gives them time to read some or all of the reviews.

In the system of equations (4) and (5) we allow the two error terms $\epsilon_i^R$ and $\epsilon_i^P$ to be correlated and estimate the system using seemingly unrelated regression (Srivastava and Giles 2020). To test a positive mediation effect for emotion $j$ we require that $\gamma_{1j} > 0$, $\beta_1 > 0$ and the indirect effect $\gamma_{1j} \times \beta_1 > 0$. Similarly, to test a negative mediation effect for emotion $j$ we require that $\gamma_{1j} < 0$, $\beta_1 > 0$ and the indirect effect $\gamma_{1j} \times \beta_1 < 0$.[13] We include the vector of emotions in equation (5) to permit a test of mediation. If $\gamma_{1j} \times \beta_1 > 0$ or $\gamma_{1j} \times \beta_1 < 0$ and $\beta_{6j} = 0$ then the average rating fully mediates the path from emotion $j$ to purchase, otherwise it partially mediates the pathway.

*6.3 Results*

The findings reveal that emotions expressed in customer reviews are often strongly associated with product evaluations and downstream purchase behavior. As expected, negative emotions in reviews correspond to lower product ratings, while positive emotions correspond to higher ratings (Rating Model), which in turn are linked to subsequent purchases by other customers (Purchase Model) (Ludwig et al., 2013; Rocklage and Fazio, 2020). Beyond these general insights, we discuss the effects of specific consumer emotions in detail next.

---

[13] To calculate standard errors for indirect effects, we use the delta method and the covariance matrix for the parameter estimates. We also calculate bootstrap confidence intervals for the indirect effects.

## 6.4 Discussion

Table 9 contains the parameter estimates for the mediation model. For the Rating model (Equation 4), consistent with prior research that links review affect, valence, and sentiment to ratings, virality, sales, and conversion rates (Chevalier and Mayzlin, 2006; Ludwig et al., 2013; Rocklage and Fazio, 2020), we find that emotions expressed in customer reviews have a significant association with product evaluations and downstream purchase behavior by other prospective customers. By using LX, we thus extend prior efforts to pinpoint which consumer emotions are linked to such effects.

**Table 9.** Purchase Model with Product Rating as a Mediator

|  | Rating Model | | | | Purchase Model | | | |
|---|---|---|---|---|---|---|---|---|
|  | **Estimate** | **SE** | **t-Stat** | ***p*-Value** | **Estimate** | **SE** | **t-Stat** | ***p*-Value** |
| **Covariates** | | | | | | | | |
| Average rating | - | - | - |  | **0.00472** | 0.00135 | 3.49 | 0.0005 |
| Price | - | - | - |  | **-0.00504** | 0.00065 | -7.81 | <.0001 |
| Volume | - | - | - |  | **0.00002** | 0.00000 | 3.13 | 0.0018 |
| Views | - | - | - |  | **0.00314** | 0.00072 | 4.36 | <.0001 |
| Review length | - | - | - |  | **0.00487** | 0.00174 | 2.80 | 0.0051 |
| **Consumer Emotions** | | | | | | | | |
| Anger | **-0.48189** | 0.04218 | -11.42 | <.0001 | 0.00387 | 0.00586 | 0.66 | 0.5092 |
| Discontent | **-0.81714** | 0.03204 | -25.50 | <.0001 | **-0.01789** | 0.00470 | -3.81 | 0.0001 |
| Worry | **0.10400** | 0.04200 | 2.48 | 0.0133 | -0.00534 | 0.00583 | -0.92 | 0.3600 |
| Sadness | **-0.34045** | 0.04890 | -6.96 | <.0001 | 0.00698 | 0.00674 | 1.04 | 0.3002 |
| Fear | **-0.14958** | 0.05589 | -2.68 | 0.0074 | 0.00655 | 0.00769 | 0.85 | 0.3941 |
| Shame | -0.07682 | 0.05148 | -1.49 | 0.1357 | -0.00784 | 0.00709 | -1.11 | 0.2688 |
| Envy | 0.08264 | 0.07705 | 1.07 | 0.2835 | -0.00916 | 0.01060 | -0.86 | 0.3875 |
| Loneliness | **0.07584** | 0.02486 | 3.05 | 0.0023 | -0.00040 | 0.00352 | -0.11 | 0.9106 |
| Romantic love | **-0.12878** | 0.04022 | -3.20 | 0.0014 | -0.00131 | 0.00554 | -0.24 | 0.8128 |
| Love | **0.15673** | 0.03406 | 4.60 | <.0001 | 0.00740 | 0.00470 | 1.57 | 0.1153 |
| Peacefulness | **-0.11657** | 0.02590 | -4.50 | <.0001 | **-0.00934** | 0.00361 | -2.59 | 0.0097 |
| Contentment | **0.52498** | 0.03139 | 16.73 | <.0001 | -0.00265 | 0.00441 | -0.6 | 0.5482 |
| Optimism | -0.00415 | 0.03174 | -0.13 | 0.896 | -0.00586 | 0.00442 | -1.33 | 0.1848 |
| Joy | **0.19309** | 0.02456 | 7.86 | <.0001 | 0.00042 | 0.00340 | 0.12 | 0.9028 |
| Excitement | **0.12571** | 0.03249 | 3.87 | 0.0001 | 0.00113 | 0.00448 | 0.25 | 0.8014 |
| Surprise | 0.03350 | 0.03650 | 0.92 | 0.3588 | -0.00924 | 0.00502 | -1.84 | 0.0657 |

Notably, discontent, anger, and sadness emerge as especially damaging to product ratings; any sign of unhappiness in reviews appears to be a red flag for other consumers. This finding aligns with evidence that negative review content can strongly damage product attitudes and purchase intentions (Chevalier and Mayzlin, 2006; Ludwig et al., 2013; Varga and Albuquerque 2024). Yet some emotions with negative valence (worry and loneliness) have significant positive associations with ratings, which initially seems counterintuitive. On reflection and linked to the focal product categories, this finding may suggest that reviewers mention their sense of worry or loneliness to describe how products (e.g., comforters, pillows) help resolve those concerns. A technology customer might note that they "were worried about buying this device, but it worked perfectly." In such cases, the emotion words ("worry," "lonely") do not signal dissatisfaction but instead highlight the positive outcome delivered by the product, resulting in a stronger positive rating. This example underscores the need to acknowledge and anticipate that the nuanced effects of specific emotions are not limited to their valence; the context and appraisal also matter (Berger and Milkman, 2012).

Joy, contentment, love, and excitement in a review exhibit significant, positive associations with star ratings, a pattern that is in line with marketing evidence that positive affect improves customer evaluations and behavioral intentions (Bagozzi, Gopinath, and Nyer 1999). Two emotions that are not conventionally identified as negative, romantic love and peacefulness, indicate significant negative relationships with product ratings. Arguably, overtly sentimental or tranquil language in a review might signal to readers that the reviewer is not sufficiently focused on product performance. Again, this evidence highlights that discrete emotions offer unique information, and positive outcomes can sometimes be communicated through ostensibly negative feelings, or vice versa.

With the two-stage modeling approach, we next use the Purchase model (Equation 5) to discern if and how the LX-detected emotions can predict actual purchases by review readers. This assessment affirms that a higher average rating is associated with purchase likelihood ($\hat{\beta}_1 = .00472$, $p = 0.0005$), consistent with substantial evidence that better ratings encourage sales (Babić-Rosario et al. 2016; Chevalier and Mayzlin, 2006; Liu, Lee, and Srinivasan 2019). The control variables also behave as expected and are in line with prior research (e.g., Liu, Lee, and Srinivasan 2019; Vana and Lambrecht 2021). For example, price has a significant negative effect on purchase (pricier products are less likely to be bought after reading reviews), whereas products with more reviews (volume), more views (popularity), and longer review length are purchased more often. Having established the link from ratings to sales, we consider how each emotion influences purchases through this mediator and offer clear evidence of mediation. That is, consumer emotions expressed in reviews indirectly drive purchase behavior through their impact on the product's star ratings.

For most emotions, the data support either a negative (negative emotions) or a positive (positive emotions) indirect association with purchase likelihood. Whereas emotions like anger or sadness reduce average ratings, which then lowers the probability of a subsequent purchase by review readers, emotions like joy or contentment elevate ratings, which increase purchases by readers. For example, Table 9 shows that anger exerts a significantly negative, indirect effect on purchase through ratings, because on the anger → rating path, it has a significant negative effect ($\hat{\gamma}_{11} = -.48189$, $p < .0001$), and then the positive rating → purchase path reveals a positive association of the average product rating with purchase behavior ($\hat{\beta}_1 = .00472$, $p = 0.0005$). If we also calculate the indirect effect of anger on purchase behavior through product rating, using the estimated index of mediation ($\hat{\gamma}_{11} \times \hat{\beta}_1$), the results establish a significantly negative, indirect effect, equal to $-.00227$ (SE $= .00068$, $p < .0008$). As recommended by Parker, Nouri, and Hayes (2011), we calculate a 95% confidence interval (CI) for the indirect effect based on 1000 bootstrap samples. The CI of [-.00385, -.00102] does not contain 0. Nor is anger directly related to purchase behavior ($\hat{\beta}_1 = .00387, p = .51$). Thus, the average product rating fully mediates the negative relationship between anger and purchase behavior. Similar patterns arise for other negative emotions such as discontent and sadness (and, to a lesser degree, fear, romantic love, and peacefulness).

On the positive side, emotions like love, contentment, joy, and excitement, as well as loneliness, exert significantly positive, indirect effects on purchases through ratings: They boost ratings and thereby encourage more purchases. From a theoretical standpoint, this behavioral validation indicates that affective review content can shape real economic outcomes (Ludwig et al. 2013; Vana and Lambrecht 2021). Beyond linking review valence or sentiment to sales, our results pinpoint which emotions are associated with these effects and also establish the predictive power of fine-grained, LX-identified emotions for aggregate consumer behavior.

However, two notable exceptions feature partial mediation, namely, discontent and peacefulness. These emotions affect ratings, but they also exhibit a significant direct effect on purchase behavior even after we control for these ratings. For discontent, the direct coefficient is negative ($\hat{\beta}_{6,2} = -.01789$, p = .0001), which implies that when prospective customers read reviews signaling discontent, they are less inclined to buy, beyond the effects of the star rating. There appears to be something uniquely discouraging about discontented language in reviews, which might not be captured by the numeric rating. This insight accords with arguments that review text provides information beyond star averages and can influence consumer decisions independently (Ludwig et al. 2013). For example, even if a product earns a respectable 4-star average, when the review content expresses a lot of discontent, consumers might hesitate, suspicious of unresolved issues.

The result for peacefulness is perhaps even more surprising; it has a small but significant negative direct effect on purchase. We speculate that the reason might be analogous to our discussion of romantic language, such that a review emphasizing peacefulness might appear irrelevant or "too emotional" to consumers who seek concrete product information. Emotional content that mismatches product or category norms can undermine credibility (Rocklage and Fazio 2020), such that if shoppers encounter a review that describes an appliance or some other utilitarian product made the reviewer feel peaceful, they might question the reviewer's objectivity or ability to evaluate functional aspects. Such skepticism could directly suppress purchase likelihood, regardless of the star rating. Because these emotional cues sway the reader, we find partial mediation, such that the emotion affects sales both through ratings and via a direct path.

## 7. General Discussion

This research addresses a central methodological challenge in marketing: the lack of validated, scalable tools for measuring consumer perceptions expressed in unstructured text in ways that align with established marketing scales. Marketers increasingly seek to use consumer-authored content, such as reviews and social media posts, but existing analytical tools tend to reduce such content to broad sentiment scores, or general emotion categories, overlooking discrete marketing relevant emotions and evaluations (Cohen, Pham, and Andrade 2018; Gopinath, Thomas, Krishnamurthi 2014; Gupta and Zeithaml 2006; Kranzbühler et al. 2020; Morgan and Hunt 1994; Palmatier et al. 2006). Dictionary-based methods lack emotional granularity (Hutto and Gilbert 2014; Pennebaker et al. 2015); generic transformer-based models mostly have not been adapted to detect validated marketing constructs (Chintagunta, Hanssens, and Hauser 2016; Netzer et al. 2012; Puranam, Kadiyali, and Narayan 2021). Therefore, we propose the Linguistic eXtractor (LX), derived by fine-tuning Meta's Llama 3.2-3B Instruct LLM on consumer-authored text, together with their self-reported perceptions measured using established marketing scales. Because this fine-tuned LX tool can detect 16 discrete consumption-related emotions (Richins 1997) and four core consumer evaluations (Morgan and Hunt 1994; Palmatier et al. 2006), it enables accurate, theory-consistent extraction of marketing-relevant constructs from unstructured text data sources, with broad research and practical implications.

## *7.1 Research Implications*

Across three studies, we assess LX's performance in relation to three central research questions: whether a fine-tuned LLM can reliably detect marketing-relevant perceptions defined by validated consumer self-reports, whether it outperforms existing tools in accuracy and generalizability, and whether LX-extracted perceptions can predict downstream marketing outcomes. Study 1 demonstrates that LX accurately identifies emotions and evaluations in open-ended survey responses and outperforms transformer and proprietary LLM baselines, including GPT-4 Turbo, DeepSeek, and RoBERTa, achieving 81% macro-F1 accuracy across 20 classification tasks. Study 2 confirms LX's generalizability using third-party-annotated reviews from Amazon and Yelp, for which it offers greater than 95% F1 accuracy. Study 3 evaluates the nomological validity of LX outputs using clickstream and purchase data from a U.K. online retailer and details how emotions extracted from customer reviews predict product ratings, which in turn predict purchase behavior. Adding to consumer emotion research (Bagozzi, Gopinath, and Nyer 1999; Puccinelli, et al. 2016), we find most emotion effects are mediated by ratings, though discontent and peacefulness also directly associated with purchase behavior, indicating that emotions in text are linked to decision-making beyond star scores alone.

These findings advance marketing theory and methods in several ways. For example, they extend literature on consumer emotions and evaluations (Bagozzi, Gopinath, and Nyer 1999; Richins 1997) by introducing a scalable, validated method for detecting fine-grained perceptions from consumer-authored text. This study enables behavioral validation of text-derived consumer emotion constructs at a scale rarely observed in prior marketing research.

Earlier work typically involved sample sizes in the hundreds and relied on self-report data (Palmatier et al. 2006), limiting their ability to generalize to marketplace behavior. Here, we analyze more than 10,000 products and thousands of product reviews and consumer browsing sessions, linking specific emotions in reviewer text to actual outcomes including ratings and purchases. This illustrates not only the predictive utility of LX, but also adds to research on the behavioral consequences of affect in consumer-generated content (Berger and Milkman 2012; Ludwig et al. 2013) by offering more fine-grained insights into how emotions influence real-world decision-making contexts. We also add to prior text analytics research (Netzer et al. 2012; Liu, Lee, and Srinivasan 2019) by demonstrating that when LLMs are fine-tuned on self-reported data, they outperform general purpose models.

*7.2 Practical Implications*

The LX model contributes to academic research and applied marketing by offering a validated, theory-aligned tool for extracting consumer perceptions from unstructured text.[14] Unlike proprietary models such as GPT-4, LX is open-source, cost-free, and privacy-secure. Fine-tuned on self-reported marketing constructs, it produces outputs that align closely with theoretical frameworks central to marketing research. The no-code web application allows users to upload consumer-authored text files (e.g., *.csv), select relevant constructs, and receive batch-level output, detailing the presence and polarity of emotions and evaluations. It thus expands access to advanced text analytics by reducing the technical and financial barriers that arise from conventional, commercial LLMs.

The LX web app translates raw consumer-authored text into numeric emotion and evaluation scores and dummies in four simple steps (Figure 3). First, users prepare a *.csv file that contains one column with a unique ID for the file to be analyzed (e.g., review, survey ID) and one column with the text to be analyzed (e.g., reviews, survey responses). Users should remove any unnecessary fields to streamline processing, but no coding or installation is required, and the app runs in any modern browser. Second, users upload the file, select the text column from a drop-down menu, and preview several rows to confirm the correct field has been chosen. Third, users pick specific consumer emotion and consumer evaluation constructs to analyze. Although LX can detect up to 16 consumption-related emotions and four consumer evaluations central to marketing, users should select the emotions or evaluations most relevant to their research context or managerial question. Fourth, users submit a brief contact form and, when processing is complete, receive a secure link to download a structured *.csv file in which each row corresponds to the original text, and columns report binary indicators (1/0) for the presence of each emotion, ±1/0 polarity for evaluations, and a word-count field. The data can be merged with any behavioral or survey data and analyzed with standard statistical packages. All processed data are deleted, so academic and managerial researchers are guaranteed privacy. For practitioners, the output supports monitoring and diagnosis at scale (e.g., flagging discontent spikes, tracking trust gains). Because the LX constructs are theory-aligned and validated, marketing researchers can use these metrics confidently to investigate relevant research topics (e.g., linking emotions to ratings and purchase).

---

[14] Reviewers may test the LX app at https://anonymouspaperdemo.vercel.app/demo. To ensure anonymity, please use a shared Gmail login (reviewerreviewer59@gmail.com, password: ReviewLX!) or non-identifiable email address. A sample data set (test.csv) is available here. No browsing data are tracked, and uploaded files must be less than 1 MB. Full details on how to use LX are available in Web Appendix F.

**Figure 3.** LX Web App Workflow: From Text to Emotion and Evaluation Scores

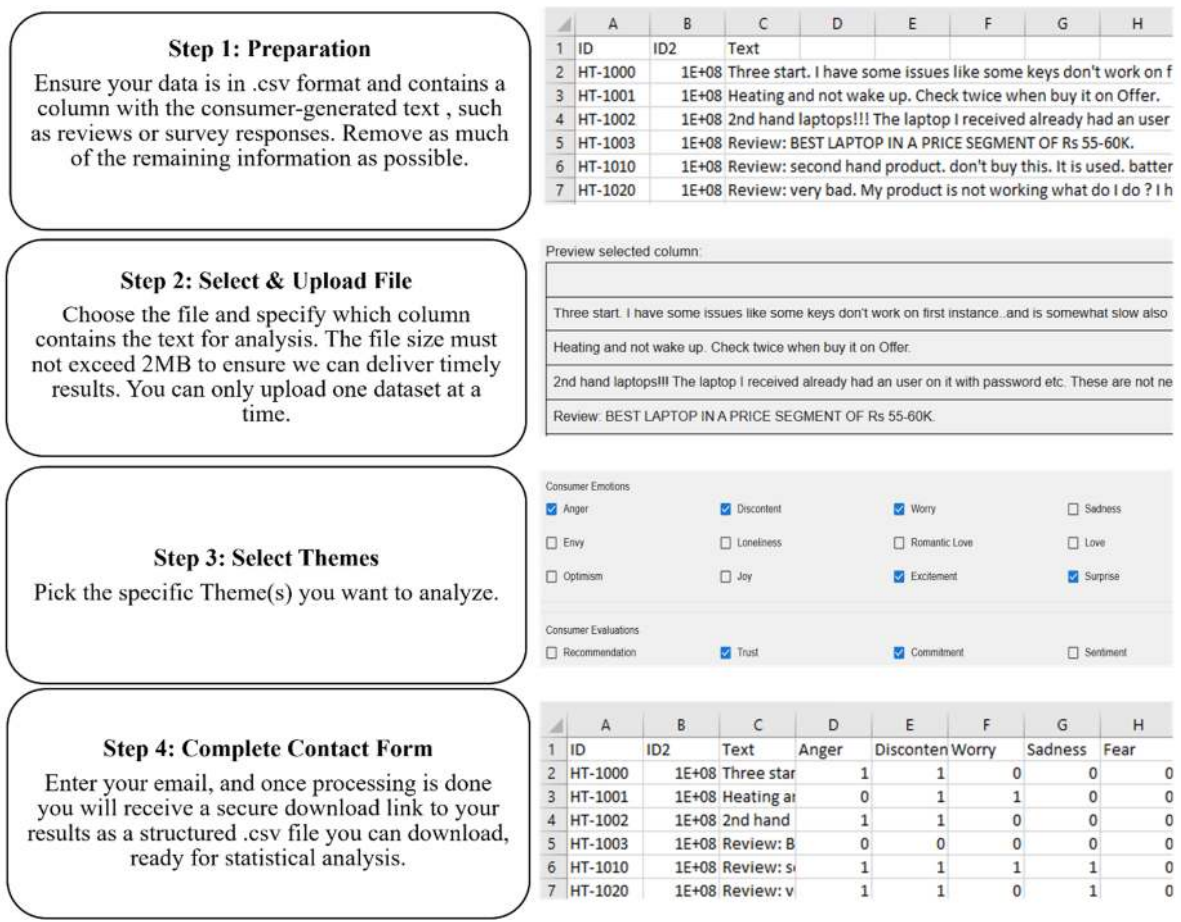

For managers, we demonstrate that the emotional tone of consumer-authored reviews—gauged according to specific emotions rather than general sentiment—provides reliable signals about product evaluations and downstream purchase behavior. Reviews expressing negative emotions such as anger and discontent are associated with lower ratings and reduced purchase likelihood; those conveying joy, contentment, love, and excitement correspond to higher ratings and increased conversion. These effects are not uniform though. Most emotional signals operate through product ratings, but discontent and peacefulness exert direct influences on purchase decisions, suggesting that emotional language conveys meaning beyond visible rating scores. LX generates scalable, targeted insights from unstructured text data, that will allow them to better track shifts in emotions. By quantifying discrete emotions and evaluations, it supports more nuanced assessments of consumer perceptions than aggregate sentiment metrics can provide.

## *7.3 Limitations and Further Research*

In introducing and validating a new tool for extracting consumer perceptions from unstructured text, this research also faces some reasonable limitations. First, the meaning and impact of emotional expressions can vary across product categories and contexts; peacefulness or excitement are likely to evoke different meanings in wellness contexts than they do in the retail reviews we study for example. We hope continued research will delineate the effects of specific emotions, according to domains, platforms, or brand positioning. Second, additional applications of LX, such as to longitudinal data (e.g., customer reviews over time), post-crisis brand commentary, or repeat consumption contexts, could help reveal how emotions evolve and whether the shifts predict consumers' behavior. Third, we train and validate the model with English-language data. Although LX's architecture is theoretically grounded, and the Llama 3 LLM supports multilingual input, the generalizability of its outputs across languages has yet to be tested. Fourth, we explicitly developed LX to analyze consumer-authored text, but marketers increasingly rely on multimodal and firm-generated content too, in their efforts to monitor sentiment and engage customers. Building on recent developments for using LLMs for visual data (e.g., Sisodia, Burnap, and Kumar 2024), continued research might develop apps to mine consumer perceptions in other data modalities (e.g., audio, imagery). Ongoing uses of the new LX web app are likely to reveal novel opportunities for further investigations too.

# Extracting Consumer Insight from Text: A Large Language Model Approach to Emotion and Evaluation Measurement

# WEB APPENDIX

# Web Appendix A. Prior Research on Consumer-Related Perceptions

*Marketing Research into Consumer-Related Perceptions*

*Data Collection and Search Strategy*

To examine the prevalence of research into key consumer-related perceptions in top marketing journals, we conducted a systematic search using the EBSCOhost database. The objective was to identify all articles published in six major marketing journals (*Journal of Marketing Research, Marketing Science, Journal of Marketing, Journal of Consumer Research, Journal of the Academy of Marketing Science,* and the *International Journal of Research in Marketing*) for the 25-year period from January 1, 2000, to December 31, 2024, that explicitly mention topics related to consumer perceptions, attitudes, and behaviors.

This search was structured around five core perceptions: emotion(s), trust, commitment, recommendation, and satisfaction/sentiment. Each perception was identified using a predefined set of search terms, systematically applied to titles, abstracts, and keywords in EBSCOhost to ensure comprehensive identification of relevant articles. After retrieving the articles, we screened them to confirm relevance and assigned them to one or more of the predefined perceptions.

*Results*

To analyze trends in the study of consumer-related perceptions, we aggregated data at two levels:
1. Annual trend analysis, reflecting the number of articles mentioning each perception, summarized by publication year. Table A1 contains a tabular overview of the number of articles for each perception per year, such that it summarizes overall trends.
- Comparative perception analysis, a more detailed breakdown that illustrates the relative prevalence of different perceptions over time. Figure 1 in the main text displays the frequency of articles that study each perception over time (2000–2024).

The full data set of articles is available at
https://osf.io/csdkr/?view_only=3ee452e4321b45a6b08f11b9799dbe32.

**Table A1.** Research into Consumer Emotions and Evaluations (2000–2024)

| Year | Emotion(s) | Trust | Commitment | Recommend | Sentiment/Satisfaction |
|------|------------|-------|------------|-----------|------------------------|
| 2000 | 1  | 3  | 10 | 3 | 19 |
| 2001 | 2  | 6  | 3  | 3 | 12 |
| 2002 | 3  | 3  | 5  | 0 | 14 |
| 2003 | 3  | 6  | 9  | 1 | 18 |
| 2004 | 4  | 1  | 8  | 3 | 16 |
| 2005 | 9  | 7  | 5  | 2 | 22 |
| 2006 | 6  | 2  | 8  | 3 | 20 |
| 2007 | 7  | 4  | 7  | 3 | 22 |
| 2008 | 5  | 9  | 5  | 1 | 17 |
| 2009 | 11 | 5  | 9  | 6 | 20 |
| 2010 | 9  | 7  | 7  | 2 | 21 |
| 2011 | 12 | 4  | 8  | 3 | 18 |
| 2012 | 7  | 8  | 4  | 1 | 17 |
| 2013 | 6  | 6  | 8  | 5 | 24 |
| 2014 | 11 | 5  | 5  | 3 | 22 |
| 2015 | 15 | 3  | 1  | 4 | 17 |
| 2016 | 5  | 6  | 5  | 1 | 25 |
| 2017 | 10 | 5  | 4  | 1 | 15 |
| 2018 | 4  | 2  | 5  | 3 | 16 |
| 2019 | 8  | 7  | 6  | 4 | 14 |
| 2020 | 11 | 8  | 7  | 3 | 16 |
| 2021 | 15 | 8  | 0  | 4 | 22 |
| 2022 | 3  | 7  | 0  | 5 | 20 |
| 2023 | 5  | 5  | 3  | 7 | 13 |
| 2024 | 8  | 10 | 7  | 3 | 18 |

*Marketing Research into Consumer Emotions and Evaluations*

For this search, we retrieved articles from the same six marketing journals, again using EBSCOhost, that contained at least one of the following terms in its title, abstract, or keywords: "text min*" OR "text analy*" OR "natural language processing" OR "NLP" OR "machine learn*" OR "topic model*" OR "sentiment analy*". The asterisks represent wildcards that capture any variations of a term, such as plural forms. Figure A1 depicts the yearly distribution of the 109 articles that used some form of text analysis. Table A2 disaggregates this set by perceptions. The number of articles is growing over time. Moreover, LLMs have not been used thus far to detect the predominant constructs investigated in marketing research (e.g., emotions and sentiment/satisfaction).

**Figure A1.** Yearly Number of Articles Using Text Analysis

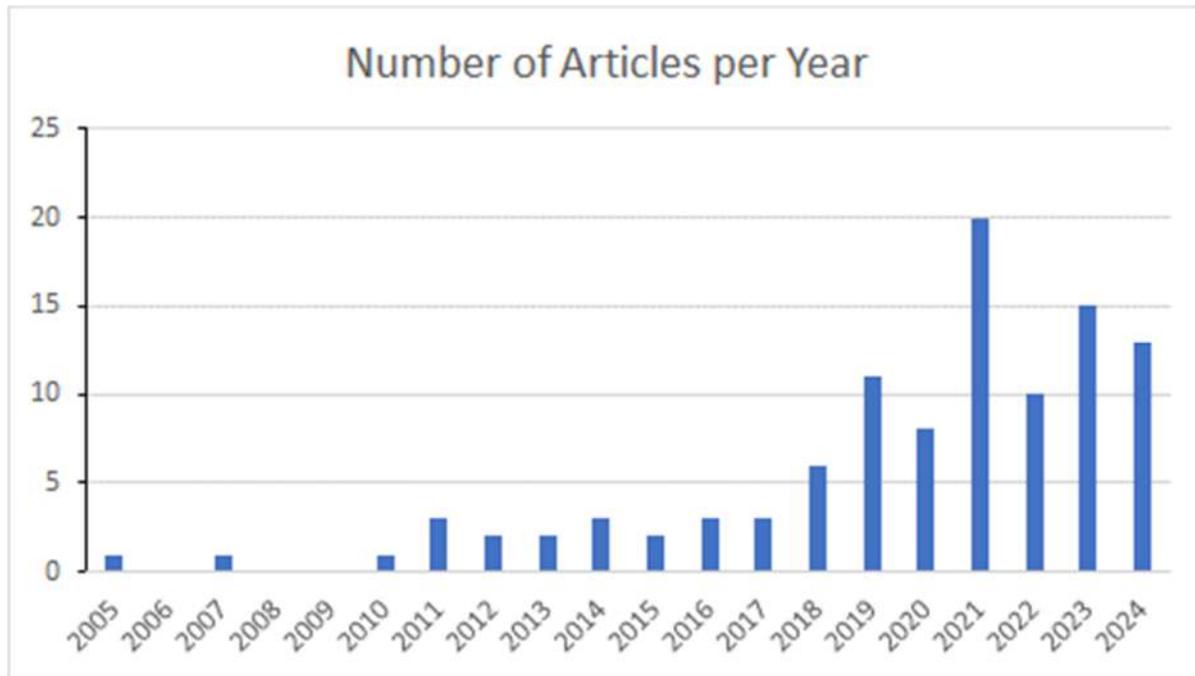

**Table A2.** Overview of Research in Marketing that Text Mines for Consumer Emotions and Evaluations

| *Reference* | *Perception* | *Context* | *Text Mining Method* |
|---|---|---|---|
| Berger and Milkman (2012) | Sentiment | Online content (*New York Times* articles) | LIWC for positive and negative sentiment |
| Berger, Kim, and Meyer (2021) | Emotions and sentiment | Movies and articles content | Developed a new dictionary: Words like laughter, happiness, and love are rated as highly positive while words like terrorist, suicide, and murder are rated as highly negative |
| Berger, Moe, and Schweidel (2023) | Sentiment | Publishers, outlets like CNBC, Wall Street Journal, Minnesota Star Tribune, and Jezebel | LIWC for positive and negative sentiment; LIWC for emotions: anger anxiety and sadness |
| Brandes and Dover (2022) | Emotions and sentiment | Hospitality/hotel | LIWC-extensions for positive and negative sentiment Anger and sadness emotions |
| Chakraborty, Kim, and Sudhir (2022) | Sentiment | Yelp restaurant review | Machine learning model (deep learning, embedding, convolutional–LSTM) |
| Herhausen et al. (2019) | Emotions and sentiment | Social media | Existing LIWC dictionaries for fear/anxiety, anger, and sadness; new dictionary for disgust |
| Herhausen et al. (2023) | Emotions and sentiment | Social media | For sentiment, existing dictionaries capture relatively negative and positive words (percentage of matching words in the complaint, e.g., "furious," "outrage"); for empathy, new |

|  |  |  | dictionary that measures all words that indicate an employee's effort to connect emotionally with the complainant by expressing understanding of their emotional state, using words of validation and affirmation. |
|---|---|---|---|
| Hovy, Melumad, and Inman (2021) | Emotions and sentiment | Social media and consumer reviews | New tool (Wordify) developed with a hybrid approach |
| Lafrenier, Moore, and Fisher (2022) | Emotions | Services and products on Yelp and Amazon | Updated LIWC's swear word dictionary of 53 word stems to (1) exclude words that do not function as swear words in the review and (2) include some swear words not in the dictionary (e.g., frick, f*ck) |
| Lee (2021) | Emotions | Social media | Supplement evaluative lexicon and leveraged LIWC affect dictionary |
| Liu, Singh, and Srinivasan (2016) | Sentiment | Tweets, IMDB reviews, and Huffington Post news | Dictionary based using LingPipe linguistic analysis package |
| Ludwig et al. (2013) | Sentiment | Book reviews on Amazon | Affective content: positive and negative |
| Luri, Schau, and Ghosh (2024) | Sentiment | Marketplace | Developed Metaphor-Enabled Marketplace Sentiment Analysis |
| Melumad, Inman, and Pham (2019) | Sentiment | Online reviews (TripAdvisor) | Negative and positive emotions from LIWC |
| Melumad and Meyer (2020) | Sentiment | Social media (Twitter) and TripAdvisor restaurant reviews | Negative emotions from LIWC |
| Mosley, Schweidel, and Zhang (2023) | Emotions and sentiment | Social media | LIWC affect dictionary, LIWC positive and negative sentiment, NRC emotion lexicon to assess the presence of anger emotion |
| Motyka et al. (2018) | Emotions | Amazon product reviews | LIWC affect dictionary |
| Mukherjee, Dutta, and De Bruyn (2022) | Emotions and sentiment | Social media | R's syuzhet package based on NRC emotion lexicon |
| Nguyen, Johnson, and Tsiros (2024) | Emotions | Emails | Deep learning, universal emotion detector (UED), EMOP (SentenceBERT, tree-based boosting algorithms) |
| Packard and Berger (2017) | NPS: Recommendation | Consumer book reviews | Search String: any word starting with the string "recomm" (e.g., "recommend," "recommendation," "recommended") or synonyms (e.g., "endorse," "suggest") |
| Rocklage et al (2023) | Sentiment and emotions | Beer Advocate | Developed a new dictionary for sentiment analysis focusing on certainty, also considering positive, negative, and emotionality |

| Roelen-Blasberg, Habel, and Klarmann (2023) | Sentiment | Products (Amazon.com, Walmart.com (beeradvocate.com and ratebeer.com) | Aspect-based sentiment analysis: integrate methods to group synonyms and co-occurrences into meaningful attributes |
| --- | --- | --- | --- |
| Sonnier, McAlister, and Rutz (2011) | Sentiment | Online communication data for a firm | Supervised machine learning model for sentiment analysis |
| Van Laer et al. (2019) | Sentiment | Online reviews (TripAdvisor) | LIWC positive and negative sentiment |
| Villarroel Ordenes et al. (2017) | Sentiment | Online customer review sites (Amazon.com, BN.com, TripAdvisor.com) and social media (Facebook and Twitter) | Enhanced LIWC dictionaries such as emoticons from PCNet. Also, Stanford Sentence and Grammatical Dependency Parser |
| Wang et al. (2022) | Sentiment | Survey data, Consumer Reports magazine, Epinions, Amazon, eBay, and news proxies for transaction data | Sentiment analysis using machine learning model, word2vec |
| Woolley and Sharif (2021) | Sentiment | Online reviews | LIWC positive and negative sentiment |

# Web Appendix B. Respondent Profile, Survey Questions, and Scales

In this appendix we provide details on the survey respondent profile and the survey questions

*Survey Respondent Profile*

Tables B1 and B2 profile the demographic composition of the 2,523 survey respondents, showing a slight gender skew toward women and a strong skew toward younger respondents. One of the screening criteria is that the respondent has previously posted a review on one of the major review platforms, so this younger profile is expected.

**Table B1.** Gender Profile for Survey Respondents

| Gender | Percentage of Respondents |
|---|---|
| Male | 42.0 |
| Female | 56.6 |
| Non-binary/third gender | 1.1 |
| Prefer not to say | 0.3 |

**Table B2.** Age Profile for Survey Respondents

| Age Range (years) | Percentage of Respondents |
|---|---|
| 18-30 | 31.9 |
| 31-40 | 30.0 |
| 41-50 | 18.5 |
| 51-60 | 12.6 |
| 61-70 | 5.3 |
| 70+ | 1.7 |

Table B3 lists the percentage of respondents who chose each of the nine categories as the focal product or service. Because the experience had to occur in the past month, we note a reasonable predominance of the retail, clothing/footwear, and restaurant categories.

**Table B3.** Product/Service Categories for Survey Respondents

| Category | Percentage of Respondents |
|---|---|
| Retail | 21.6 |
| Clothing/Footwear | 15.6 |
| Furnishings | 5.0 |
| Restaurant | 27.1 |
| Health | 6.3 |
| Transport | 6.3 |
| Communication | 5.2 |
| Recreation | 9.8 |
| Insurance | 3.2 |

*Emotions and Their Definition*

We refer to Richins's (1997) 16 emotions, as listed and defined in Table B4.

**Table B4.** Definitions of Emotions Captured in the Survey Data

| Emotion Label | Emotion Items/Descriptors (Richins 1997) |
|---|---|
| Anger | Frustrated, angry, irritated |
| Discontent | Unfulfilled, discontented |
| Worry | Nervous, worried, tense |
| Sadness | Depressed, sad, miserable |
| Fear | Scared, afraid, panicky |
| Shame | Embarrassed, ashamed, humiliated |
| Envy | Envious, jealous |
| Loneliness | Lonely, homesick |
| Romantic love | Sexy, romantic, passionate |
| Love | Loving, sentimental, warm hearted |
| Peacefulness | Calm, peaceful |
| Contentment | Contented, fulfilled |
| Optimism | Optimistic, encouraged, hopeful |
| Joy | Happy, pleased, joyful |
| Excitement | Excited, thrilled, enthusiastic |
| Surprise | Surprised, amazed, astonished |

*Emotion and Evaluation Profile of Survey Respondents*

Table B5 reveals the number and percentage of times respondents mentioned each of the 16 emotions. The most prevalent negative emotions are anger and discontent; the most prevalent positive emotions are contentment, joy, and peacefulness. On average, respondents mentioned 2.9 of the 16 emotions while writing about the emotions they felt when using the product/service.

**Table B5.** Profile of Emotions Mentioned in the Survey Data

| Emotion Label | Number of Mentions | Percentage of Mentions |
|---|---|---|
| Anger | 1003 | 15.8 |
| Discontent | 902 | 14.2 |
| Worry | 393 | 6.2 |
| Sadness | 302 | 4.8 |
| Fear | 151 | 2.4 |
| Shame | 158 | 2.5 |
| Envy | 30 | 0.5 |
| Loneliness | 39 | 0.6 |
| Romantic love | 54 | 0.9 |
| Love | 301 | 4.8 |
| Peacefulness | 650 | 10.3 |
| Contentment | 982 | 15.5 |
| Optimism | 537 | 8.5 |
| Joy | 939 | 14.8 |
| Excitement | 628 | 9.9 |
| Surprise | 266 | 4.2 |
|  |  |  |
| Total | 6332 | 100.0 |

Then Table B6 indicates the average number of words written for each perception and the median time that respondents spend writing about each perception. The overall average number of words is greater than 30 words, with not much variation across perceptions. The median writing time is close to 1 minute for each perception, and the entire survey took about 14 minutes. We asked respondents to write a minimum of 15 words, and they needed to spend no longer than 2 minutes per perception. Thus, respondents substantially exceeded the minimum length requirement, and they spent sufficient time on each perception, which suggests the written tasks were undertaken thoughtfully.

**Table B6.** Average Word Count for Written Reviews for each Perception

| Perception | Average Written Text Word Count | Median Text Writing Time (seconds) |
|---|---|---|
| Consumer Emotions | 32.7 | 54.4 |
| Trust | 33.5 | 58.1 |
| Commitment | 33.6 | 55.0 |
| Recommendation | 33.0 | 54.2 |
| Consumer sentiment for price | 30.4 | 46.5 |
| Consumer sentiment for product | 33.8 | 58.4 |
| Consumer sentiment for place | 30.6 | 50.8 |
| Consumer sentiment for communication | 32.2 | 55.1 |
|  |  |  |
| Total Across Survey | 291.2 | 854.0 |

# Web Appendix C. Illustration of Probability Updating to Minimize Loss

As explained in the main text, a common criterion for LLM fine-tuning is the cross-entropy function:

$$\mathcal{L} = -\sum_{i=1}^{K} \log\left(\Pr\left(y_i^{(t)} | y_{<i}^{(t)}, (I^{(t)}, x), W\right)\right), \tag{C1}$$

where $\Pr\left(y_i^{(t)} | y_{<i}^{(t)}, (I^{(t)}, x), W\right)$ is the conditional probability for the next token in the output sequence, given the input data, instructions, output thus far, and current model parameters $W$. These probabilities are determined by the LLM on the basis of the full vocabulary (corpus), with probabilities calculated using the softmax function, which is similar to multinomial logit functions frequently used in marketing literature.

For example, if a pretrained LLM vocabulary consisted of five words (joy, anger, discontent, not, present), these five words allow for the identification of three emotions, being present or not present. Table C1 is a schematic of the true output sequence for the online review in Table 1 in the main text, along with the vocabulary and conditional probability of each word in the vocabulary being the next token in the output sequence.

**Table C1.** Pretrained Probabilities for Emotion Detection for Output Sequence "joy present"

| True labels | joy | present | |
|---|---|---|---|
| Vocab word index | Position 1 probability | Position 2 probability | Vocabulary |
| 1 | 0.30 | 0.05 | 1 joy |
| 2 | 0.25 | 0.05 | 2 anger |
| 3 | 0.20 | 0.05 | 3 discontent |
| 4 | 0.20 | 0.40 | 4 not |
| 5 | 0.05 | 0.45 | 5 present |

For position 1 in the output sequence, the true token is *joy*, and it has the highest probability of any word in the vocabulary (.30). Using greedy sampling, the LLM selects the token with the highest probability (*joy* in this case). In this pretrained LLM, there is not much difference in the probabilities for the three emotions in the vocabulary. For the second token in the output sequence, there also is not much difference between the probabilities for *not* and *present*, but the LLM choses *present* due to its slightly higher probability. Using Equation C1, the loss function for the pretrained LLM is $\mathcal{L} = -\log(0.3) - \log(0.45) = 2.002$.

After fine-tuning the LLM using data from the survey, in which consumers wrote about experiences that might elicit joy, this emotion indeed gets revealed as one of the emotions. In turn, the LLM's parameters are updated, as are the probabilities for each token in the vocabulary. Table C2 gives the revised probabilities for the online review in Table 1.

**Table C2.** Fine-Tuned Probabilities for Emotion Detection, Output Sequence "joy present"

| True labels | joy | present | |
|---|---|---|---|
| Vocab word index | Position 1 probability | Position 2 probability | Vocabulary |
| 1 | 0.55 | 0.05 | 1 joy |
| 2 | 0.20 | 0.05 | 2 anger |
| 3 | 0.15 | 0.05 | 3 discontent |
| 4 | 0.05 | 0.15 | 4 not |
| 5 | 0.05 | 0.80 | 5 present |

The loss function for the fine-tuned LLM is now $\mathcal{L} = -\log(0.55) - \log(0.80) = 0.8210$, which is less than the loss function value of 2.002 for the pretrained LLM. This illustration outlines the basic principles by which LLMs update their conditional probabilities after being fine-tuned. It relies on prompts and data to update LLM parameters and thus minimize the loss function across all training data.

# Web Appendix D. Training and Test Data

As mentioned in the main paper, for the perceptions of marketing evaluations and aspect-based sentiment, the survey data lack sufficient quantities of the neutral category, so they need to be supplemented with online review data from Amazon and Yelp for the neutral outcome category. We aim for an approximate balance of negative, positive, and neutral written texts in the training and test data. Table D1 contains the profile of outcome categories for each perception in the training data; as it shows, approximately equal numbers of text reviews constitute the negative, positive, and neutral categories. The neutral category further can be divided into text sourced from survey and review data. The supplementary review data clearly help expand the number of written texts in the neutral category.

**Table D1.** Output Profile for Each Perception for Training Data

| Perception | Number Negative | Number Positive | Number Neutral | | |
|---|---|---|---|---|---|
| | Survey Data | Survey Data | All Data | Survey Data | Review Data |
| Trust | 545 | 531 | 537 | 66 | 471 |
| Commitment | 516 | 545 | 528 | 57 | 471 |
| Recommendation | 556 | 542 | 542 | 504 | 38 |
| Consumer sentiment for price | 479 | 503 | 497 | 47 | 450 |
| Consumer sentiment for product | 599 | 590 | 604 | 62 | 542 |
| Consumer sentiment for place | 393 | 383 | 405 | 65 | 340 |
| Consumer sentiment for communication | 662 | 681 | 664 | 93 | 571 |
| | | | | | |
| Total Across Test Data | 3750 | 3775 | 3777 | 894 | 2883 |

Table D2 provides the profile of outcome categories for each perception for the test data. Again, we see an approximate balance across negative, positive, and neutral categories. Review data often are used to supplement the neutral category for all perceptions except recommendation, which mirrors the training data observations.

**Table D2.** Output Profile for Each Perception for Test Data

| Perception | Number Negative | Number Positive | Number Neutral | | |
|---|---|---|---|---|---|
| | Survey Data | Survey Data | All Data | Survey Data | Review Data |
| | | | | | |
| Trust | 127 | 141 | 135 | 18 | 117 |
| Commitment | 146 | 117 | 134 | 14 | 120 |
| Recommendation | 127 | 141 | 141 | 128 | 13 |
| Consumer sentiment for price | 137 | 113 | 119 | 12 | 107 |
| Consumer sentiment for product | 148 | 156 | 143 | 9 | 134 |
| Consumer sentiment for place | 99 | 109 | 87 | 10 | 77 |
| Consumer sentiment for communication | 174 | 154 | 172 | 29 | 143 |
| | | | | | |
| Total Across Test Data | 958 | 931 | 931 | 220 | 711 |

# Web Appendix E. Details on Comparison Models

In domains outside marketing literature, several transformer models and LLMs have been applied for emotion detection and sentiment analysis (e.g., Feng et al. 2024; Peng et al. 2023, Wake et al. 2023). We present several state-of-the-art models that have the capability to undertake the research tasks and that offer a high prediction standard:

- BART[1]: Lewis et al. (2019) developed the BART transformer model for text classification. When used as a zero-shot classifier (classification without further training), given an input text and a target perception, the output categories are constructed as candidate labels. The BART model then classifies the input text to the candidate label with the highest predicted probability.
- RoBERTa[2]: Another transformer model, it can be used in the same manner as BART for zero-shot classification.
- Llama 3.2-8B-Instruct[3]: This version of the Llama model contains 8 billion parameters and is trained to follow instructions, in the same way as the baseline model, Llama3.2-3B-Instruct.
- GPT 3.5-turbo-instruct[4]: This closed-source model, provided by OpenAI, has an estimated 20 billion parameters (OpenAI does not officially disclose the number).[5] We use this model in the same way as Llama3.2-3B-Instruct. It imposes costs of $1.50 per 1 million input tokens and $2.00 per 1 million output tokens.[6]
- GPT 4-Turbo: This closed-source model provided by OpenAI has an estimated 1.76 trillion parameters. We use it in the same way as Llama3.2-3B-Instruct and GPT-3.5-turbo-instruct. The costs are $10.00 per 1 million input tokens and $30.00 per 1 million output tokens.[7]
- Deep Seek -V3: A recently released open-source LLM with 671 billion parameters, it is pretrained on 14.8 trillion tokens.[8] The costs for using this model are $0.27 per 1 million input tokens and $1.10 per 1 million output tokens.

The transformer models (BART and RoBERTa) are zero-shot, meaning they receive a prompt without any example being supplied. For the LLMs (Llama-3.2-3B-Instruct, Llama3.2-8B-Instruct, GPT 3.5-turbo-instruct, GPT 4-Turbo, Deep Seek-V3), we use only pretrained versions. The results we present thus support comparisons of the pretrained Llama-3.2-3B-Instruct with a fine-tuned version (namely, LX), which in turn help demonstrate the benefits of our fine-tuning approach. Because GPT 3.5-turbo-instruct and GPT 4-Turbo are closed-sourced LLMs, they cannot be fine-tuned in the same way (for an explanation, see Ye, Yoganarasimhan, and Zheng 2025). Thus, we use only the pretrained versions of these OpenAI LLMs.

---

[1] https://huggingface.co/facebook/bart-large-mnli
[2] https://huggingface.co/FacebookAI/roberta-large-mnli
[3] https://huggingface.co/meta-llama/Llama-3.1-8B-Instruct
[4] https://blog.nextideatech.com/openai-gpt-3-5-turbo-instruct/
[5] https://community.openai.com/t/how-many-parameters-does-gpt-3-5-have/648417/3
[6] https://openai.com/api/pricing/
[7] https://artificialanalysis.ai/models/gpt-4-turbo
[8] https://huggingface.co/deepseek-ai/DeepSeek-V3

# Web Appendix F. LX User Manual

**NOTE TO THE REVIEW TEAM:** The LX Web Application and User Manual are provided by the author team. For the review process, the actual LX website is not shown. Instead, we provide access to an anonymized, single-page demonstration site that includes the key operational steps. To preserve anonymity, we have removed all identifying information from the demonstration website. If the paper is accepted, we will update this user manual to reflect the final public version of the application.

Preserving user anonymity is tricky; we have taken several precautions to ensure that you can experience the app without revealing either your or our identities. Access the demo site at [**https://anonymouspaperdemo.vercel.app/demo**]. It functions similarly to the final version, though the interface has been simplified. Your browsing behavior is not tracked, so feel free to explore the application, upload a *.csv file, and analyze your own data.

To further ensure your anonymity, we recommend using an email address that is not personally identifiable. Alternatively, you may log in using the following shared Gmail account that we have created only for the purpose of reviewing:
  **Email:** reviewerreviewer59@gmail.com
  **Password:** ReviewLX!

This account allows you to submit your data using an OTP and receive the processed results without revealing your identity. Please note that for testing purposes, uploaded files should not exceed 1 MB in size. If you do not have your own consumer text data, you are welcome to use our sample file, available for download here: https://github.com/SAP-samples/datahub-dine/blob/main/data/reviews.csv. We have no affiliation with the provider of these reviews, so you can download this review.csv file by clicking on the download button on the right-hand side of the screen. We have tested it, and there are no viruses or malware associated with this file.

1. **Before you start**

The **LX Web Application** is a browser-based tool for analyzing consumer-authored text that describes products or services using a fine-tuned large language model (LLM). It is optimized to detect marketing-relevant **emotions** (e.g., joy, optimism, discontent, anger) and **evaluations** (e.g., trust, commitment, sentiment) and thereby support contemporary research in marketing and consumer psychology. No installation is required; LX runs entirely online and is compatible with all major browsers including Chrome, Firefox, Safari, and Edge. All processing is conducted on our server once you have submitted the data. All data are deleted immediately after the analysis, ensuring privacy and security. No programming knowledge is needed.
  Access the demonstration site at: [**https://anonymouspaperdemo.vercel.app/demo**]
2. **Prepare Your Data**
  Create a *.csv file with a column containing all consumer-authored text. Each row should represent one entry (e.g., review, survey response, comment). The column header should be clearly labeled (e.g., "text"), as in Figure F1.

**Figure F1.** Example *.csv File Format

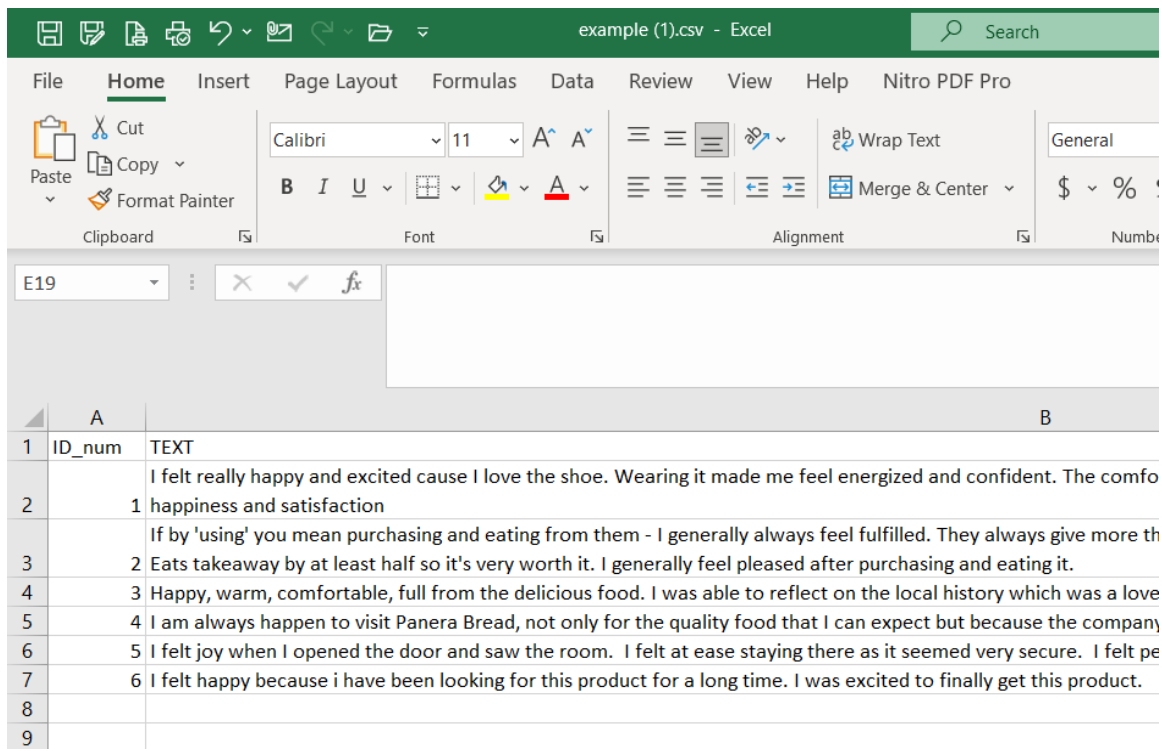

## 3. Access the Web App

Navigate to [**https://anonymouspaperdemo.vercel.app/demo**], verify that you are a human, and accept the Terms and Conditions by reviewing these and clicking "I agree" (see Figure F2).

**Figure F2**: Screenshot of Landing Page

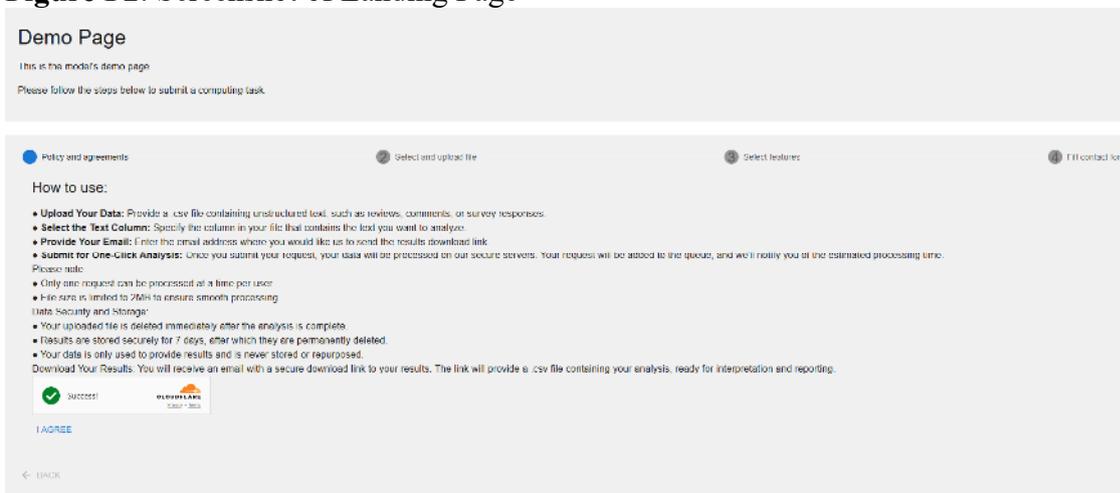

## 4. Upload Your File

Click **"Choose file"** and select your *.csv file (see Figure F3). Only standard *.csv UTF-8 encoded files are accepted.

**Figure F3**: Screenshot of File Upload Interface

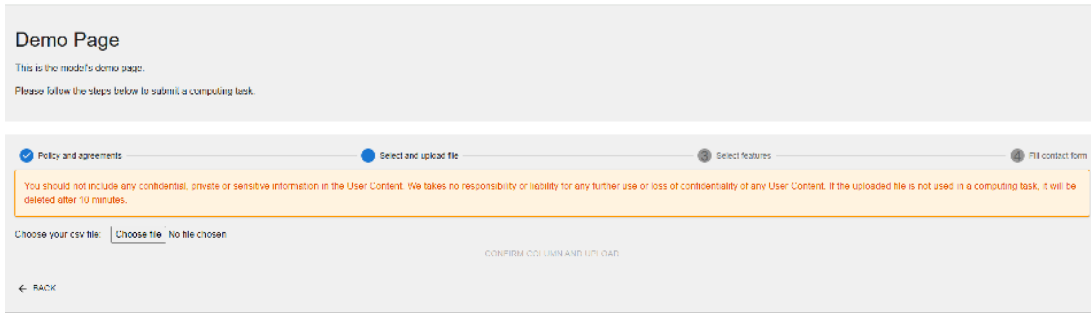

## 5. Select the Text Column

Use the dropdown menu to choose the column containing the text you want analyzed, as in Figure F4.

**Figure F4**: Screenshot of Column Selection Step

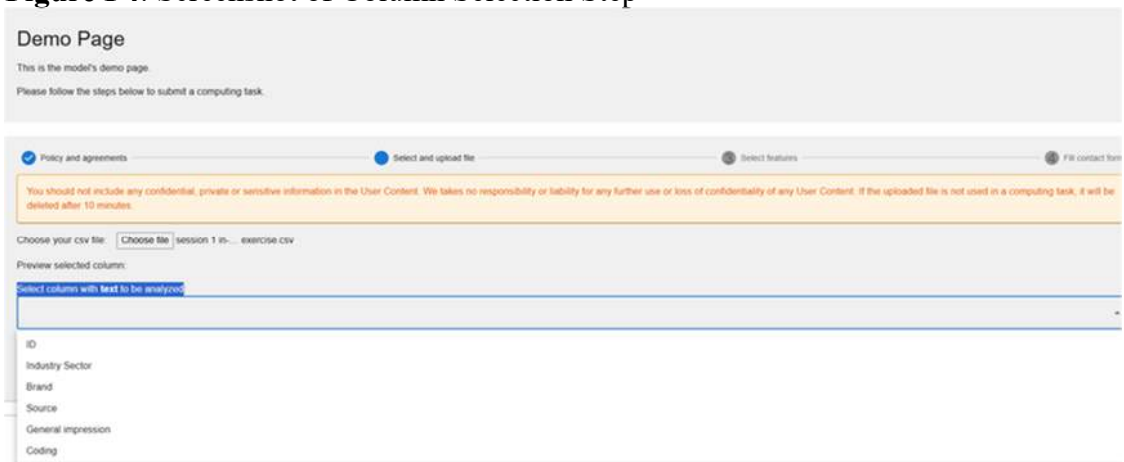

## 6. Preview the Text Column, Confirm and Upload
Check that the selected column contains the correct text using the preview feature (see Figure F5). Once satisfied, confirm and upload your data.

**Figure F5**: Screenshot of Column Preview and Confirmation

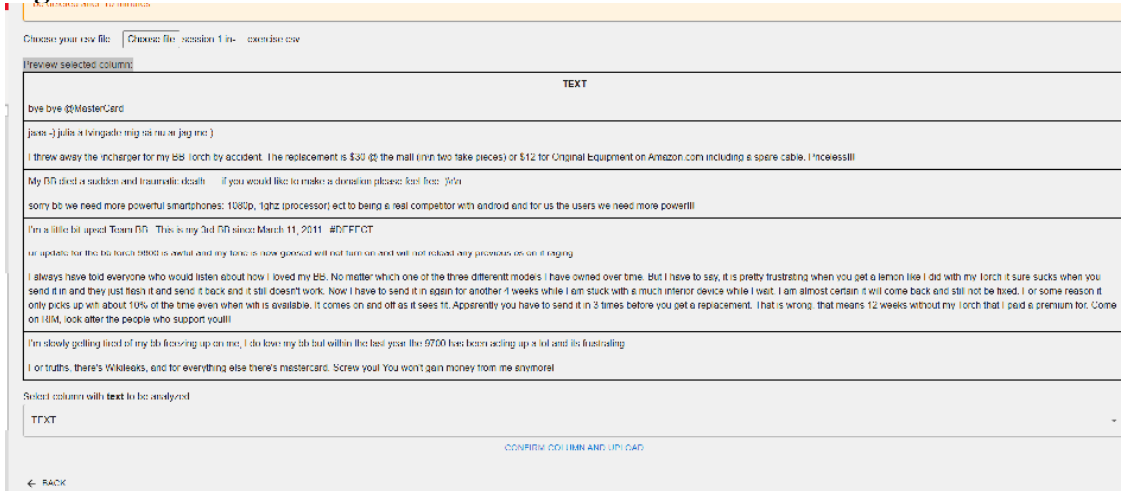

7. **Select Analysis Perceptions**

Choose the **emotion** and evaluation perceptions you want the model to text mine. Then click **Confirm**. In some cases, you may want to analyze only a subset of perceptions, and LX allows you to selectively choose which categories of emotions and evaluations to analyze. For example, if your data set focuses on brand trust or sentiment, you only need to select these specific constructs. To do so, check the boxes next to the perceptions you wish to include in your analysis (Figure F6). If you want the model to analyze all available perceptions, click the "Select All" button. Note that LX supports analyses across the following perception groups:

**Consumer Emotions** *(select any of the following)*
- Anger
- Discontent
- Worry
- Sadness
- Fear
- Shame
- Envy
- Loneliness
- Romantic Love
- Love
- Peacefulness
- Contentment
- Optimism
- Joy
- Excitement
- Surprise

**Consumer Evaluations** *(select any of the following))*
- Trust
- Commitment
- Recommendation
- Sentiment

Only selected perceptions will be included in the analysis output. Your selections will remain in effect during the current text mining task. This flexibility enables you to tailor the output to the specific perceptions you believe are present or relevant in your consumer text data set. If you do not believe that a specific perception is discussed in your data or relevant for your research question, we recommend that you not mine for it.

**Figure F6**: Screenshot of Perception Selection

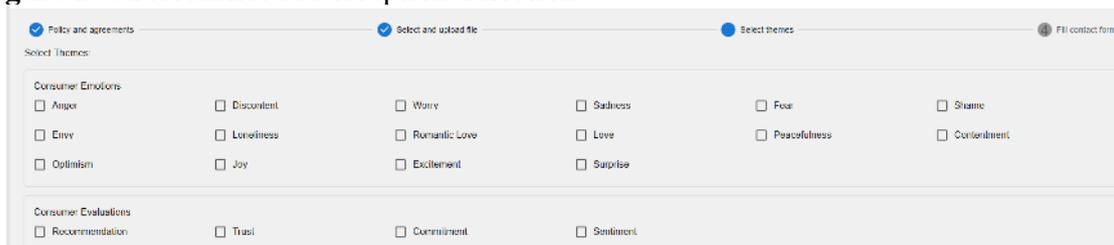

8. **Complete the Contact Form**

Enter an email address where the results will be sent (Figure F7). Click **"Send OTP"** to receive a 6-digit verification code. Paste the code into the form to confirm. [**NOTE TO REVIEWERS: Please leave the Name and Affiliation fields blank**.] Press **Submit**.

**Figure F7**: Screenshot of Contact Form

## 9. Submission Complete

Once your file has been uploaded successfully, the site will display **"All steps completed – you are finished."** You will receive an email with an estimated processing time.

## 10. Download Your Results

When processing is complete, you will receive another email containing a secure download link. Your output *.csv file will include the original text, along with binary indicators for which emotions were present in each text and any evaluation polarities (see Figure F8). The *.csv file can be imported into most statistical packages (e.g., SPSS, R, SAS). The LX output variables are given as dummies (1 = present, 0 = not present) for consumer emotions; polarities for all evaluations (1 = positive, -1 = negative, 0 = no mention or neutral polarity); or total word counts in each text cell. For details on the output, please see Table 4 in the main text. Results are stored for 7 days, after which all files are permanently deleted.

**Figure F8**: Screenshot of Output, CSV Format

# References Used in the Web Appendix

Gerard Pfann (2013), "More than Words: The Influence of Affective Content and Linguistic Style Matches in Online Reviews on Conversion Rates," *Journal of Marketing*, 77 (1), 87–103.

Luri, Ignacio, Hope J. Schau, and Bikram Ghosh (2024), "Metaphor-enabled marketplace sentiment analysis," *Journal of Marketing Research,* 61 (3), 496-516.

Melumad Shiri, J. Jeffrey Inman, and Michel Pham (2019), "Selectively emotional: How smartphone use changes user-generated content," *Journal of Marketing Research*, 56 (2), 259-275.

Melumad Shiri and Robert Meyer (2020), "Full disclosure: How smartphones enhance consumer self-disclosure," *Journal of Marketing*, 84 (3), 28-45.

Mosley Buffy, David A. Schweidel, and Kunpeng Zhang (2024), "When connection turns to anger: How consumer–brand relationship and crisis type moderate language on social media," *Journal of Consumer Research*, 50 (5), 907-922.

Motyka Scott, Dhruv Grewal, Elizabeth Aguirre, Dominik Mahr, Ko De Ruyter, and Martin Wetzels (2018), "The emotional review–reward effect: how do reviews increase impulsivity?" *Journal of the Academy of Marketing Science* 46 (6), 1032-1051.

Mukherjee Prithwiraj, Souvik Dutta S, and Arnaud De Bruyn (2022), "Did clickbait crack the code on virality?" *Journal of the Academy of Marketing Science*, 50 (3), 482-502.

Nguyen Nguyen, Joseph Johnson, and Michael Tsiros (2024), "Unlimited testing: Let's test your emails with AI," *Marketing Science*, 43 (2), 419-439.

Packard Grant and Jonah Berger (2017), "How language shapes word of mouth's impact," *Journal of Marketing Research* 54 (4), 572-588.

Peng, Liyizhe, Zixing Zhang, Tao Pang, Jing Han, Huan Zhao, Hao Chen, and Björn W. Schuller (2023), "Customising General Large Language Models for Specialised Emotion Recognition Tasks," arXiv: https://arxiv.org/abs/2310.14225.

Richins, Marsha L (1997), "Measuring emotions in the consumption experience," *Journal of Consumer Research,* 24 (2), 127-146.

Rocklage Matthew D., Dharlene He, Derej D. Rucker, and Loran F. Nordgren (2023), "Beyond Sentiment: The Value and Measurement of Consumer Certainty in Language," *Journal of Marketing Research*, 60 (5), 870-888.

Roelen-Blasberg Tobias, Johannes Habel, and Martin Klarmann (2023), "Automated inference of product attributes and their importance from user-generated content: Can we replace traditional market research?" *International Journal of Research in Marketing*, 40 (1), 164-188.

Sonnier Garrett P., Leigh McAlister, and Olver J, Rutz (2011), "A dynamic model of the effect of online communications on firm sales," *Marketing Science*, 30 (4), 702-716.